\documentclass{vllm}

\usepackage[margin=1.2in]{geometry}
\usepackage{amsmath,amssymb,amsthm}
\usepackage{mathtools}
\usepackage{tikz}
\usetikzlibrary{shapes.geometric,arrows.meta,positioning,fit,backgrounds,
                calc,decorations.pathreplacing,matrix,shadows,patterns}
\usepackage{pgfplots}
\pgfplotsset{compat=1.18}
\usepackage{xcolor}
\usepackage{booktabs}
\usepackage{microtype}
\usepackage{hyperref}
\usepackage{cleveref}
\usepackage{enumitem}
\usepackage{caption}
\usepackage{subcaption}
\usepackage{cite}
\usepackage{graphicx}
\usepackage{float}
\usepackage{multirow}
\usepackage{tabularx}
\usepackage{colortbl}

\theoremstyle{plain}
\newtheorem{thesis}{Thesis}
\newtheorem{observation}{Observation}
\theoremstyle{definition}
\newtheorem{definition}{Definition}
\newtheorem{opportunity}{Open Opportunity}
\crefname{opportunity}{Opportunity}{Opportunities}
\Crefname{opportunity}{Opportunity}{Opportunities}

\newenvironment{frisk}{%
  \par\vspace{6pt}%
  \noindent{\color{routergreen!30}\rule{\linewidth}{0.4pt}}%
  \par\vspace{3pt}\noindent
  {\sffamily\bfseries\color{routergreen!70!black}%
   Feasibility and risk.\enspace}\begingroup\small\ignorespaces
}{\endgroup\par}

\newcommand{\eg}{\emph{e.g.},\ }

\newcommand{\etal}{\emph{et al.}}

\newcommand{\vsr}{\textsc{vLLM-SR}}

\newcommand{\boundary}{\mathcal{B}}
\DeclareMathOperator{\tokW}{tok/W}

\definecolor{workloadblue}{HTML}{2171B5}
\definecolor{routergreen}{HTML}{238B45}
\definecolor{poolorange}{HTML}{D94801}
\definecolor{objectivepurple}{HTML}{6A3D9A}
\definecolor{lightgray}{HTML}{F0F0F0}
\definecolor{ownwork}{HTML}{C6534D}

\title{\textbf{The Workload--Router--Pool Architecture for LLM Inference Optimization}\\[4pt]
       {\large A Vision Paper from the vLLM Semantic Router Project}}

\author{%
  Huamin Chen$^{1}$ \quad
  Xunzhuo Liu$^{1}$ \quad
  Bowei He$^{2}$ \quad
  Fuyuan Lyu$^{3,4}$ \quad
  Yankai Chen$^{2}$ \\[0.4em]
  Xue Liu$^{1,2,3,4}$ \quad
  Yuhan Liu$^{5}$ \quad
  Junchen Jiang$^{6}$
}

\affiliation[1]{vLLM Semantic Router Project}
\affiliation[2]{MBZUAI}
\affiliation[3]{McGill University}
\affiliation[4]{Mila} 
\affiliation[5]{University of Chicago}
\affiliation[6]{Tensormesh Inc.\ /\ UChicago}

\date{March 2026}

\abstract{%
Over the past year, the vLLM Semantic Router project has released a
series of work spanning:
(1)~\emph{core routing mechanisms}---signal-driven routing,
context-length pool routing, router performance engineering, policy
conflict detection, low-latency embedding models, category-aware
semantic caching, user-feedback-driven routing adaptation,
hallucination detection, and hierarchical content-safety
classification for privacy and jailbreak protection;
(2)~\emph{fleet optimization}---fleet provisioning and
energy-efficiency analysis;
(3)~\emph{agentic and multimodal routing}---multimodal agent routing,
tool selection, CUA security, and multi-turn context memory and
safety;
(4)~\emph{governance and standards}---inference routing protocols and
multi-provider API extensions.
Each paper tackled a specific problem in LLM inference, but the
problems are not independent; for example, fleet provisioning depends
on the routing policy, which depends on the workload mix, shifting as
organizations adopt agentic and multimodal workloads.
This paper distills those results into the
\emph{Workload--Router--Pool (WRP) architecture}, a
three-dimensional framework for LLM inference optimization.
\textbf{Workload} characterizes what the fleet serves (chat
vs.\ agent, single-turn vs.\ multi-turn, warm vs.\ cold,
prefill-heavy vs.\ decode-heavy).
\textbf{Router} determines how each request is dispatched (static
semantic rules, online bandit adaptation, RL-based model selection,
quality-aware cascading).  \textbf{Pool} defines where inference
runs (homogeneous vs.\ heterogeneous GPU, disaggregated
prefill/decode, KV-cache topology).  We map our prior work
onto a $3 \times 3$ WRP interaction matrix, identify which cells we
have covered and which remain open, and propose twenty-one concrete research directions at the
intersections, each grounded in our prior measurements, tiered by
maturity from engineering-ready to open research.
}

\begin{document}
\maketitle

\section{Introduction}
\label{sec:intro}

Three variables shape production LLM inference: the workload, the
routing strategy, and the GPU pool architecture.  These are studied by
largely separate communities---workload characterization, model
routing, and systems/fleet optimization---that seldom interact.  But in
practice, a routing decision tuned for one workload on one pool
topology can be suboptimal for a different combination.

\paragraph{Our research program.}
The vLLM Semantic Router (\vsr{}) project~\cite{vllmsr2026} has been
building the \emph{inference routing stack}.  Across a growing body
of work (summarized in \Cref{tab:ownwork} and
\Cref{sec:foundation}), we have addressed four pillars:
\begin{itemize}[nosep]
  \item \textbf{Routing architecture}: signal-driven semantic routing
        with composable signals~\cite{vllmsr2026}, conflict-free
        policy languages for probabilistic ML
        predicates~\cite{probpol2026}, 98$\times$ router latency
        reduction via Flash Attention and prompt
        compression~\cite{fastrouter2026}, 2D Matryoshka embeddings
        for configurable latency--quality
        trade-offs~\cite{mmberted2025}, reasoning-selective
        routing~\cite{wang2025reason}, category-aware semantic
        caching~\cite{wang2025cache}, feedback-driven online
        routing adaptation~\cite{feedbackdet2026},
        hallucination-aware response validation with conditional
        factcheck classification and token-level
        detection~\cite{halugate2025, factcheck2026}, and
        hierarchical content-safety classification following the
        MLCommons AI Safety Taxonomy with a binary safe/unsafe
        gate~\cite{safetyl1_2026} and a 9-class hazard
        categorizer~\cite{safetyl2_2026} enabling per-category policy
        enforcement and privacy-preserving routing;
  \item \textbf{Pool routing and fleet optimization}: token-budget pool
        routing~\cite{poolrouting2026}, FleetOpt analytical fleet
        provisioning with gateway-layer
        compression~\cite{fleetopt2026}, inference-fleet-sim for
        queueing-grounded capacity
        planning~\cite{fleetsim2026}, and the 1/W law for energy
        efficiency~\cite{onewlaw2026};
  \item \textbf{Multimodal and agent routing}: adaptive VLM routing for
        computer use agents~\cite{avr2026}, outcome-aware tool
        selection~\cite{oats2026}, the Visual Confused Deputy
        security guardrail~\cite{vcd2026}, and the
        gateway-centric personal-assistant stack
        OpenClaw~\cite{openclaw2026} (open-source; multi-channel
        sessions and tools behind a single Gateway control plane);
  \item \textbf{Governance and standards}: the Semantic Inference
        Routing Protocol (SIRP) at IETF~\cite{sirp2025} and
        multi-provider extensions for agentic AI inference
        APIs~\cite{mpext2025}.
\end{itemize}

Each paper solved a specific problem, but the problems are coupled.  
For example, pool routing~\cite{poolrouting2026} creates a cost
cliff whose resolution via gateway-layer compression depends on the
workload's prompt-length CDF archetype~\cite{fleetopt2026}.  The 1/W
law~\cite{onewlaw2026} shows that routing topology is a stronger
energy lever than GPU generation, but only when the pool supports
context-length partitioning.

\begin{figure}[t]
      \centering
      \begin{tikzpicture}[
          >=Stealth,
          box/.style={draw, rounded corners=3pt, minimum width=3.2cm,
                      minimum height=1cm, font=\small\sffamily, align=center,
                      line width=0.8pt},
          dimbox/.style={box, line width=1.2pt, minimum width=4cm,
                         minimum height=1.3cm, font=\sffamily\bfseries},
          arr/.style={->, line width=0.8pt},
          darr/.style={<->, line width=0.8pt, dashed},
      ]
      \node[dimbox, fill=workloadblue!15, draw=workloadblue]
          (W) at (0, 0) {Workload\\Characterization};
      \node[dimbox, fill=routergreen!15, draw=routergreen]
          (R) at (5.5, 0) {Router\\Layer};
      \node[dimbox, fill=poolorange!15, draw=poolorange]
          (P) at (11, 0) {Pool\\Architecture};
      
      \node[box, fill=workloadblue!5, draw=workloadblue!60,
            minimum width=3.6cm, font=\scriptsize\sffamily]
          (w1) at (0, -2) {Chat vs.\ Agent};
      \node[box, fill=workloadblue!5, draw=workloadblue!60,
            minimum width=3.6cm, font=\scriptsize\sffamily]
          (w2) at (0, -3.2) {Warm vs.\ Cold};
      \node[box, fill=workloadblue!5, draw=workloadblue!60,
            minimum width=3.6cm, font=\scriptsize\sffamily]
          (w3) at (0, -4.4) {Prefill-heavy vs.\ Decode-heavy};
      
      \node[box, fill=routergreen!5, draw=routergreen!60,
            minimum width=3.6cm, font=\scriptsize\sffamily]
          (r1) at (5.5, -2) {Static semantic rules};
      \node[box, fill=routergreen!5, draw=routergreen!60,
            minimum width=3.6cm, font=\scriptsize\sffamily]
          (r2) at (5.5, -3.2) {Online bandit / feedback};
      \node[box, fill=routergreen!5, draw=routergreen!60,
            minimum width=3.6cm, font=\scriptsize\sffamily]
          (r3) at (5.5, -4.4) {RL / cascading selection};
      
      \node[box, fill=poolorange!5, draw=poolorange!60,
            minimum width=3.6cm, font=\scriptsize\sffamily]
          (p1) at (11, -2) {Homo vs.\ Hetero GPU};
      \node[box, fill=poolorange!5, draw=poolorange!60,
            minimum width=3.6cm, font=\scriptsize\sffamily]
          (p2) at (11, -3.2) {Disaggregated P/D};
      \node[box, fill=poolorange!5, draw=poolorange!60,
            minimum width=3.6cm, font=\scriptsize\sffamily]
          (p3) at (11, -4.4) {KV-cache topology};
      
      \foreach \s in {w1,w2,w3} \draw[arr, workloadblue!70] (W) -- (\s);
      \foreach \s in {r1,r2,r3} \draw[arr, routergreen!70] (R) -- (\s);
      \foreach \s in {p1,p2,p3} \draw[arr, poolorange!70] (P) -- (\s);
      
      \draw[darr, objectivepurple] (W) -- (R)
          node[midway, above=1pt, font=\scriptsize\itshape, text=objectivepurple]
          {workload shapes routing};
      \draw[darr, objectivepurple] (R) -- (P)
          node[midway, above=1pt, font=\scriptsize\itshape, text=objectivepurple]
          {routing shapes pool need};
      \draw[darr, objectivepurple] (W) to[bend left=25] node[midway, above=2pt,
          font=\footnotesize\itshape, text=objectivepurple] {workload determines pool sizing} (P);
      
      \definecolor{safetyred}{RGB}{180,40,60}
      \node[draw=safetyred, fill=safetyred!8, rounded corners,
            minimum width=12cm, minimum height=0.8cm,
            font=\small\sffamily\bfseries, text=safetyred]
          (safety) at (5.5, -5.8) {Safety \& Privacy: Content Gate
          \textbullet\ Jailbreak Detection
          \textbullet\ Hazard Classification
          \textbullet\ Privacy Routing};
      
      \draw[arr, safetyred!50] (safety) -- (0, -5.1);
      \draw[arr, safetyred!50] (safety) -- (5.5, -5.1);
      \draw[arr, safetyred!50] (safety) -- (11, -5.1);
      
      \node[draw=objectivepurple, fill=objectivepurple!8, rounded corners,
            minimum width=12cm, minimum height=0.8cm,
            font=\small\sffamily\bfseries, text=objectivepurple]
          (obj) at (5.5, -7.2) {Optimization Objectives: Cost
          \textbullet\ Accuracy \textbullet\ Latency \textbullet\ Energy};
      
      \draw[arr, objectivepurple!50] (obj) -- (0, -6.3);
      \draw[arr, objectivepurple!50] (obj) -- (5.5, -6.3);
      \draw[arr, objectivepurple!50] (obj) -- (11, -6.3);
      \end{tikzpicture}
      \caption{The Workload--Router--Pool (WRP) architecture.  Three
      dimensions interact through dashed bidirectional arrows.  Safety \&
      Privacy is a cross-cutting layer: content gating and hazard
      classification shape all three dimensions.  All are further
      constrained by the optimization objectives at the bottom.  Our
      prior work addresses one or more cells in this framework.}
      \label{fig:wrp}
      \end{figure}
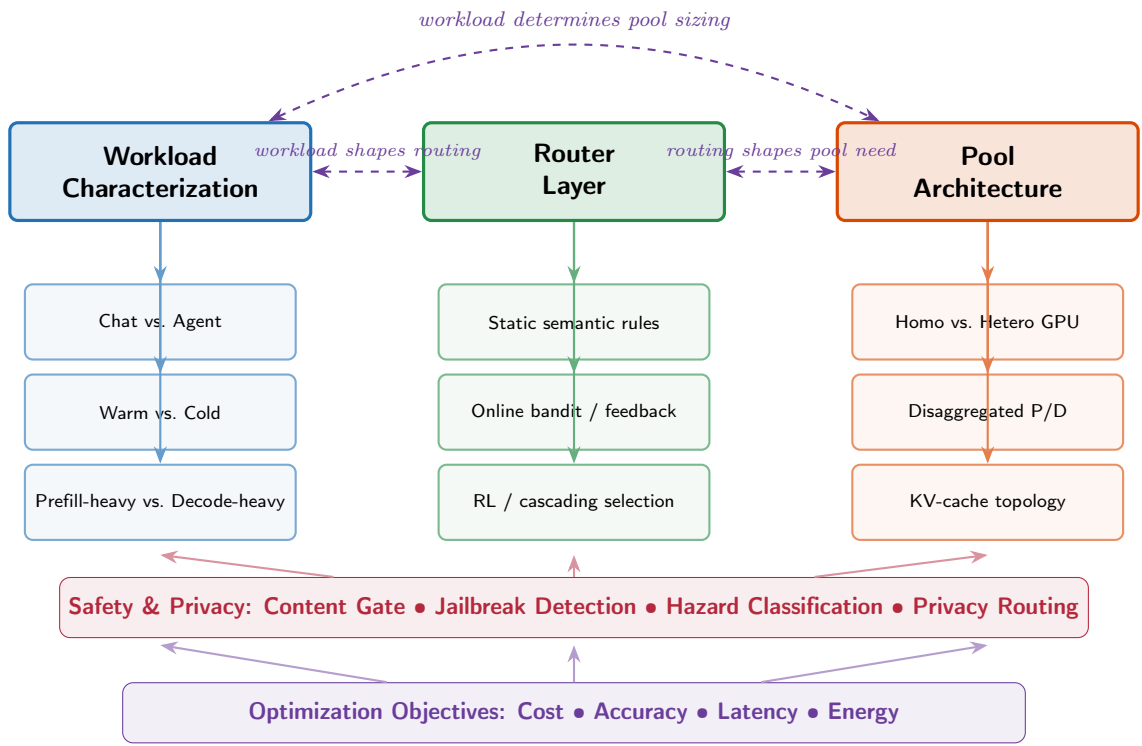

\paragraph{Our contributions.}
We present the \textbf{Workload--Router--Pool (WRP) architecture}
(\Cref{fig:wrp}), a three-dimensional framework that organizes both
our prior work and the broader research landscape.  The
contributions are:
\begin{enumerate}[nosep]
  \item A \textbf{structural decomposition} into three interacting
        dimensions---Workload, Router, Pool---grounded in evidence from
        our previous work.
  \item A \textbf{mapping} of our prior work onto the
        $3 \times 3$ WRP interaction matrix, showing which cells we
        have addressed and which remain open.
  \item A \textbf{research roadmap} of twenty-one concrete opportunities at
        the intersections, each grounded in our prior measurements.
\end{enumerate}

\section{Foundation: The vLLM Semantic Router Research Program}
\label{sec:foundation}

\Cref{tab:ownwork} lists the publications that underpin this paper.
We group them into four pillars.

\begin{table}[t]
\centering
\caption{Publications from the vLLM Semantic Router project that
form the foundation of the WRP vision.  Each row indicates which WRP
dimension(s) the paper primarily addresses.
\textbf{W}=Workload, \textbf{R}=Router, \textbf{P}=Pool.}
\label{tab:ownwork}
\small
\renewcommand{\arraystretch}{1.15}
\begin{tabularx}{\textwidth}{l X c c c}
\toprule
\textbf{Short name} & \textbf{Title / venue} & \textbf{W} & \textbf{R} & \textbf{P} \\
\midrule
\rowcolor{lightgray}
PoolRouting~\cite{poolrouting2026}
  & Token-Budget-Aware Pool Routing (arXiv) & $\bullet$ & $\bullet$ & $\bullet$ \\
FleetOpt~\cite{fleetopt2026}
  & Analytical Fleet Provisioning with C\&R (arXiv) & $\bullet$ & $\bullet$ & $\bullet$ \\
\rowcolor{lightgray}
1/W Law~\cite{onewlaw2026}
  & Context-Length Routing and Energy Efficiency (arXiv) & $\bullet$ & $\bullet$ & $\bullet$ \\
\midrule
FeedbackDet~\cite{feedbackdet2026}
  & mmBERT-32K Feedback Detector (HuggingFace) & $\bullet$ & $\bullet$ &  \\
\rowcolor{lightgray}
AVR~\cite{avr2026}
  & Adaptive VLM Routing for CUAs (arXiv) & $\bullet$ & $\bullet$ &  \\
MPExt~\cite{mpext2025}
  & Multi-Provider Extensions for Agentic AI (IETF) & $\bullet$ & $\bullet$ &  \\
\rowcolor{lightgray}
OpenClaw~\cite{openclaw2026}
  & Personal assistant + Gateway (GitHub OSS) & $\bullet$ & $\bullet$ &  \\
SemCache~\cite{wang2025cache}
  & Category-Aware Semantic Caching (arXiv) &  & $\bullet$ & $\bullet$ \\
FleetSim~\cite{fleetsim2026}
  & Queueing-Grounded Fleet Capacity Planner (arXiv) & $\bullet$ &  & $\bullet$ \\
\midrule
\rowcolor{lightgray}
\vsr{}~\cite{vllmsr2026}
  & Signal-Driven Decision Routing for MoM Models (arXiv) &  & $\bullet$ &  \\
ProbPol~\cite{probpol2026}
  & Conflict-Free Policy for Probabilistic ML Predicates (arXiv) &  & $\bullet$ &  \\
\rowcolor{lightgray}
FastRouter~\cite{fastrouter2026}
  & 98$\times$ Faster LLM Routing (arXiv) &  & $\bullet$ &  \\
WhenToReason~\cite{wang2025reason}
  & Semantic Router for vLLM (NeurIPS MLForSys) &  & $\bullet$ &  \\
\rowcolor{lightgray}
mmBERT-Embed~\cite{mmberted2025}
  & 2D Matryoshka Embeddings for Routing (HuggingFace) &  & $\bullet$ &  \\
HaluGate~\cite{halugate2025}
  & Token-Level Hallucination Detection (vLLM Blog) &  & $\bullet$ &  \\
\rowcolor{lightgray}
FactCheck~\cite{factcheck2026}
  & mmBERT-32K Factcheck Classifier (HuggingFace) &  & $\bullet$ &  \\
OATS~\cite{oats2026}
  & Outcome-Aware Tool Selection (arXiv) &  & $\bullet$ &  \\
\rowcolor{lightgray}
VCD~\cite{vcd2026}
  & Visual Confused Deputy: CUA Security (arXiv) &  & $\bullet$ &  \\
SafetyL1~\cite{safetyl1_2026}
  & MLCommons Binary Safety Classifier (HuggingFace) &  & $\bullet$ &  \\
\rowcolor{lightgray}
SafetyL2~\cite{safetyl2_2026}
  & MLCommons 9-Class Hazard Classifier (HuggingFace) &  & $\bullet$ &  \\
SIRP~\cite{sirp2025}
  & Semantic Inference Routing Protocol (IETF) &  & $\bullet$ &  \\
\bottomrule
\end{tabularx}
\end{table}

\subsection{Pillar 1: Routing Architecture}
\label{sec:foundation:routing}

The \vsr{} position paper~\cite{vllmsr2026} introduced
signal-driven decision routing: composing heterogeneous
signals---keyword patterns, embedding similarity, domain classifiers,
language detection---into deployment-specific routing policies across
thirteen signal types.  ProbPol~\cite{probpol2026}
formalized conflict detection when probabilistic ML signals co-fire,
and mmBERT-Embed~\cite{mmberted2025} provides the embedding backbone
with 2D Matryoshka flexibility for configurable
latency--quality trade-offs (\Cref{sec:router:static}).

A design constraint is that the router must be \emph{low-latency
and low-resource}: it should not require a dedicated GPU that could
otherwise serve LLM inference.
FastRouter~\cite{fastrouter2026} enforces this for long-context
requests with a 98$\times$ latency reduction and sub-800\,MB GPU
footprint (\Cref{sec:router:static}).

Beyond request-time dispatch, the routing stack includes
response-time mechanisms.
WhenToReason~\cite{wang2025reason} classifies whether a query needs
reasoning-mode invocation.
SemCache~\cite{wang2025cache} eliminates redundant inference via
category-aware semantic caching with domain-specific similarity
thresholds (\Cref{sec:workload:warm-cold}).
The Feedback Detector~\cite{feedbackdet2026} enables online routing
adaptation from per-turn user satisfaction signals
(\Cref{sec:router:online}).
HaluGate~\cite{halugate2025} and the Factcheck
Classifier~\cite{factcheck2026} extend the router to response-time
validation, steering factual queries toward reliable models and
feeding per-model hallucination rates back into routing policy
(\Cref{sec:router:static}).

The routing stack also enforces content safety and privacy
through a hierarchical classification pipeline following the
MLCommons AI Safety Taxonomy.  The Level-1 binary
classifier~\cite{safetyl1_2026} gates every inbound request as
safe or unsafe in a single forward pass; requests flagged unsafe
are escalated to the Level-2 nine-class hazard
classifier~\cite{safetyl2_2026}, which maps them to specific
categories (violent crimes, privacy, misinformation, etc).
This two-stage design keeps routing overhead low for benign
traffic while enabling per-category policy enforcement---e.g.,
blocking privacy-violating prompts outright, routing
misinformation-prone queries to models with stronger factual
grounding, or logging specialized-advice requests for compliance
auditing (\Cref{sec:router:static}).

\subsection{Pillar 2: Pool Routing and Fleet Optimization}
\label{sec:foundation:fleet}

Token-budget pool routing~\cite{poolrouting2026} was the starting
point: dispatching to short-context or long-context pools based on
estimated total token budget, reducing GPU instances by 17--39\% on
production traces (\Cref{sec:router:static}).

FleetOpt~\cite{fleetopt2026} unified pool routing with gateway-layer
compression into an analytical framework that derives the
minimum-cost two-pool fleet with optimal boundary $\boundary^*$ and
compression parameter $\gamma^*$, given a workload CDF and latency
target.  
Complementing this, inference-fleet-sim~\cite{fleetsim2026} validated these
results with a discrete-event simulator covering A10G, A100, and
H100 GPUs (\Cref{sec:pool:homo-hetero}).

The 1/W law~\cite{onewlaw2026} quantifies the energy dimension:
routing topology is a stronger energy lever than GPU generation, and
the gains compose multiplicatively (\Cref{sec:pool:energy}).

\subsection{Pillar 3: Multimodal and Agent Routing}
\label{sec:foundation:agent}

Adaptive VLM Routing (AVR)~\cite{avr2026} extended routing to
vision-language CUA tasks with a three-tier framework: Tier~1
multimodal classifier ($\sim$10\,ms pre-route), Tier~2 small-VLM
confidence probing with token-filtered scoring, Tier~3 large-model
escalation.  AVR projects up to 78\% cost reduction while staying
within 2\,pp of an all-large-model baseline.

OATS~\cite{oats2026} addressed tool selection in the critical
request path: offline interpolation of tool embeddings toward
success-query centroids improves NDCG@5 from 0.869 to 0.940 on
MetaTool without serving-time cost.

The Visual Confused Deputy~\cite{vcd2026} reframed CUA perception
errors as a security vulnerability, proposing dual-channel
contrastive classification that independently evaluates click targets
and reasoning text, achieving F1$=$0.915.  Combined with AVR, VCD provides both efficiency and safety
through one routing pipeline.

\subsection{Pillar 4: Governance and Standards}
\label{sec:foundation:governance}

At the standards level, the Semantic Inference Routing Protocol
(SIRP)~\cite{sirp2025} specifies a framework for content-level
classification and semantic routing in AI inference systems at the
IETF.  The Multi-Provider Extensions~\cite{mpext2025} extend agentic
AI inference APIs for multi-provider environments.

\section{The WRP Architecture}
\label{sec:wrp}

Our publications support a thesis:

\begin{thesis}[WRP Coupling]
\label{thesis:wrp}
The three dimensions of LLM inference optimization---\textbf{Workload},
\textbf{Router}, and \textbf{Pool}---are coupled, not orthogonal.
Optimizing any single dimension in isolation leaves significant
efficiency on the floor.  The largest gains require co-optimization
across dimensions.
\end{thesis}

Evidence from our own work: FleetOpt~\cite{fleetopt2026} demonstrates
that co-designing the router compression parameter $\gamma$ with pool
sizes $(n_s, n_l)$ yields 3.1--6.4\% lower cost than retrofitting
compression onto a pre-existing fleet (\textbf{Router $\times$ Pool}
coupling).  The 1/W law~\cite{onewlaw2026} shows that the same GPU
fleet can vary 40$\times$ in energy efficiency depending on the context
window served (\textbf{Workload $\times$ Pool} coupling).
AVR~\cite{avr2026} shows that VLM workloads require entirely different
routing signals than text-only chat (\textbf{Workload $\times$ Router}
coupling).

\begin{definition}[WRP Decomposition]
A production LLM inference deployment is characterized by a triple
$(\mathbf{W}, \mathbf{R}, \mathbf{P})$ where:
\begin{itemize}[nosep]
  \item $\mathbf{W}$: the \emph{workload} distribution over request
        types, prompt lengths, decode lengths, session structures, and
        modalities;
  \item $\mathbf{R}$: the \emph{router} that maps each request to a
        (model, pool, configuration) tuple; and
  \item $\mathbf{P}$: the \emph{pool architecture} specifying GPU
        types, pool boundaries, disaggregation topology, and KV-cache
        management.
\end{itemize}
The optimization objective is a weighted combination of \emph{cost}
(GPU-hours per request), \emph{accuracy} (task completion quality),
\emph{latency} (P99 TTFT, TPOT), and \emph{energy} (tokens per watt),
subject to SLO constraints.
\end{definition}

\section{Dimension 1: Workload Characterization}
\label{sec:workload}

Our fleet optimization work~\cite{poolrouting2026, fleetopt2026,
fleetsim2026} showed that workload characterization drives the remaining
design decisions.  We organize workloads
along four axes.

\subsection{Chat vs.\ Agent Workloads}
\label{sec:workload:chat-agent}

Chat workloads (LMSYS-Chat-1M~\cite{zheng2024lmsys},
the Azure LLM Inference Trace published alongside
Splitwise~\cite{patel2024splitwise}) are dominated by short prompts:
80--95\% of requests fall below 8K tokens, with a heavy tail to 64K+.
Agent workloads (SWE-bench~\cite{jimenez2024swebench},
BFCL~\cite{patil2025bfcl}) exhibit prompt lengths that grow
deterministically with turn number as tool outputs accumulate.
ServeGen~\cite{servegen2025} reports that agents make 3--10$\times$
more LLM calls per user request than chatbots.

Both types are \emph{multi-turn}, but with different session
structures.  Chat sessions are conversational: each turn
carries implicit context from prior exchanges, and the user may
shift topics, provide corrections, or gradually escalate adversarial
probes across turns.  Agent sessions are task-driven: context grows
monotonically as tool outputs accumulate, and the session terminates
on task completion or failure.  Multi-turn APIs (such as OpenAI's
Responses API) encode these patterns explicitly, offering
conversation-state management and turn-level optimizations that
single-request routing cannot exploit.  A multi-turn-aware router must maintain session context for two
reasons: KV-cache affinity and routing signal quality.  Prior turns
reveal user intent, topic drift, difficulty escalation, and safety
violations that a single-turn classifier would miss.

Our pool routing paper~\cite{poolrouting2026} first exposed how
the chat/agent distinction creates routing opportunities: the
80--95\% of short chat requests are over-provisioned in a
homogeneous 64K fleet, wasting $7/8$ of KV-cache slots.  The
deterministic turn-by-turn growth of agent sessions suggests that
session-aware prediction could outperform per-category EMA for
routing.

\paragraph{Memory management as a hidden cost multiplier.}
Beyond raw context growth, multi-turn sessions face a compounding
cost problem rooted in memory management failures.  When an LLM
hallucinates or selects the wrong tool, the erroneous output enters
the conversation history and is re-sent on every subsequent
turn---the organization pays input-token cost for content that
\emph{reduced} accuracy.  The agent must either tolerate the
polluted context (risking further errors through
self-conditioning~\cite{driftbench2026}) or spend additional LLM
calls to summarize, compress, or remove the bad turns before
continuing.  AgentHallu~\cite{agenthallu2026} benchmarks this
challenge: even frontier models achieve only 41.1\% accuracy at
localizing which step caused a hallucination, dropping to 11.6\%
for tool-use hallucinations---making targeted cleanup unreliable and
often requiring full-context re-summarization.

A comprehensive memory
survey~\cite{memorysurv2026} identifies five mechanism families for
agent memory (context-resident compression, retrieval-augmented
stores, reflective self-improvement, hierarchical virtual context,
and policy-learned management), each with distinct cost profiles.
A direct cost-performance comparison~\cite{memcost2026} shows that
at 100K-token context lengths, per-turn inference charges grow
proportionally with context even under prompt caching, and
fact-based memory systems become cheaper after approximately ten
interaction turns---precisely the regime where agent sessions
operate.  Context compression can mitigate the
problem: ACON~\cite{acon2025} achieves 26--54\% peak-token reduction
while preserving 95+\% task accuracy via failure-driven guideline
optimization, and Focus~\cite{focus2026} demonstrates 22.7\% token
savings through autonomous agent-driven context pruning on SWE-bench
tasks.  However, both operate on the agent side, optimizing a
single session in isolation.  This is a fundamental limitation:
SAMULE~\cite{samule2025} shows that \emph{single-trajectory}
reflection (the only level available to an individual agent)
produces narrow error corrections, while \emph{inter-task} learning
that aggregates failure patterns across diverse sessions yields
transferable insights that significantly outperform per-session
baselines.  CORRECT~\cite{correct2025} reinforces this: multi-agent
errors recur with similar structural patterns across requests, and
an online cache of distilled error schemata---built from prior
sessions---improves step-level error localization by up to 19.8\%.
Mistake Notebook Learning~\cite{mnl2025} demonstrates the same
principle: batch-clustered failure analysis produces generalizable
guidance that prevents known pitfalls without parameter updates.

The router has a structural advantage that directly exploits this
cross-session transferability: it observes outcomes from thousands
of sessions per domain across all tenants, aggregating failure
patterns into the per-\emph{(model, domain, failure-class)} table
(\Cref{opp:failure-routing}).  An individual agent, by contrast, sees
only its own session---too few examples to build reliable failure
statistics or discover domain-specific compression policies (e.g.,
``coding sessions tolerate aggressive early-turn pruning of failed
tool traces; medical sessions must retain full context for compliance
audits'').  The router provides this cross-session intelligence as a
fleet-wide service that every agent benefits from immediately,
including new agents that would otherwise face a cold-start problem.

\paragraph{Gateway-native assistants and fleet observability.}
OpenClaw~\cite{openclaw2026} is a concrete open-source example of the
pattern \emph{personal assistant + local-first Gateway}: users talk
to the agent over many messaging channels while a WebSocket Gateway
coordinates sessions, tool execution (browser, automation, device
nodes), and model calls.  Such deployments generate long, messy
multi-turn traces---tool outcomes, retries, channel switches, and
compaction events---that fixed academic benchmarks rarely capture.
The inference router sits on the hot path for model traffic from
those gateways and can log which model, tool surface, and safety path
correlated with short vs.\ runaway sessions.  Aggregating traces
across tenants supplies offline RL and fleet-prior datasets that a
single local assistant cannot
amass~(\Cref{opp:session-cascade}, \Cref{opp:tool-pruning}).
Conversely, routing policies tuned for token- and round-minimization
feed back into operators' choices of model defaults and tool budgets
at the Gateway.

\paragraph{Proxy-level instruction, tool, and memory surfaces.}
ITR~\cite{itr2026} retrieves minimal system-instruction fragments
and a narrowed tool subset per agent step; on the authors' controlled
agent benchmark it reports $\sim$95\% lower per-step context tokens
and large gains in correct tool routing because monolithic catalogs
can consume $\sim$90\% of the context window.  The same mechanism can be implemented at an
inference \emph{gateway}---rewriting \texttt{tools} payloads and
system blocks before the request reaches the model---so frameworks
that always send full catalogs still benefit from fleet-wide
calibration.  ToolScope~\cite{toolscope2025} merges and filters tools
on the agent side; SMART~\cite{smart2026} shows that curbing tool
overuse improves both cost and downstream accuracy.  A router that
coordinates narrowing, merging hints, and budget signals
(BATS~\cite{bats2025}) avoids duplicating incompatible policies in
every SDK.  Finally, memory is not only ``what the agent
remembers'' but \emph{what the next prefill pays for}: MemCost-style
analyses~\cite{memcost2026} show when fact stores beat raw context,
and session-aware schedulers (AgServe~\cite{agserve2025},
Continuum~\cite{continuum2025}) treat KV-cache lifetime and
multi-turn structure as first-class resources.  The router is the
natural place to choose among full history, injected summaries, and
retrieval-first prefill for the next turn---reducing tokens
\emph{and} the error compounding that lengthens
rounds~(\Cref{opp:gateway-agent-loops}).

\subsection{Warm vs.\ Cold Requests}
\label{sec:workload:warm-cold}

\emph{Warm} requests hit a KV-cache prefix from a prior turn or
shared system prompt; \emph{cold} requests start from scratch.
Cached tokens cost up to $10\times$ less~\cite{llmd2026kvcache},
and prefix-cache-aware scheduling achieves $57\times$ faster
response times in distributed deployments~\cite{llmd2026kvcache}.

For agentic workloads, 40--60\% of session time involves paused tool
calls, causing LRU eviction to collapse cache hit
rates~\cite{vllm2026rfc37003}.  Session-affinity routing---keeping a
session on the same pool instance---could preserve KV-cache locality.
Category-aware semantic caching~\cite{wang2025cache} complements this
at a higher level: domain-aware similarity thresholds and TTLs
eliminate redundant inference requests entirely---returning cached
responses for semantically equivalent queries without invoking the
LLM, reducing both cost and load on the serving pool.

\subsection{Prefill-Heavy vs.\ Decode-Heavy Workloads}
\label{sec:workload:prefill-decode}

Vision-language models encode images into hundreds of visual tokens,
making VLM workloads \emph{prefill-heavy}.  Our AVR
system~\cite{avr2026} confronted this directly: CUA screenshots are
$\sim$5\,MB, and routing 4K screenshots requires multimodal signal
extraction.  FastRouter~\cite{fastrouter2026} showed that classical
NLP compression can cap router input to $\sim$512 tokens regardless
of original prompt length, making prefill-heavy workloads tractable
for the routing layer itself.

Coding agents are \emph{decode-heavy}: code generation produces long
structured outputs.  The decode ratio directly affects whether
disaggregated prefill/decode architectures
(Splitwise~\cite{patel2024splitwise},
DistServe~\cite{zhong2024distserve}) provide benefit, and our
inference-fleet-sim~\cite{fleetsim2026} tool models this trade-off
explicitly.

\subsection{Workload Identity Signals}
\label{sec:workload:identity}

Our prior work treats incoming requests as opaque text and infers
workload type through content analysis---embedding similarity,
keyword matching, and domain classifiers.  In practice, the request
structure itself carries rich identity signals that the router can
exploit without content inspection, opening the door to
zero-latency workload classification.

\paragraph{Explicit API-level signals.}
Modern inference APIs already expose structural metadata that
distinguishes agent from chat workloads:
\begin{itemize}[nosep]
  \item \textbf{Tool definitions.}  A request carrying a \texttt{tools}
        array with function schemas is almost certainly an agent
        workload; the number and types of tools (code execution,
        web search, MCP connectors) further classify the agent's
        operational mode.  The tool catalog itself is a cost lever:
        ITR~\cite{itr2026} shows that dynamically exposing only the
        minimal tool subset per step reduces per-step context tokens
        by 95\% and improves correct tool routing by 32\%, because
        monolithic tool catalogs consume up to 90\% of available
        context.
  \item \textbf{Session chaining.}  Fields such as
        \texttt{previous\_response\_id} (OpenAI Responses API) or
        conversation-state objects signal multi-turn continuations.
        Their presence tells the router the request is mid-session
        (enabling KV-cache retention, \Cref{opp:kv-retention}) and
        provides the turn number without parsing conversation history.
  \item \textbf{System-prompt fingerprints.}  System prompts are
        typically static per deployment and can be hashed at the
        gateway.  A lookup table of known hashes maps each deployment
        to its domain, expected tool set, and historical session-length
        distribution---giving the router a per-request prior at
        zero classification cost.
  \item \textbf{Gateway metadata headers.}  Production gateways
        (Cloudflare AI Gateway's \texttt{cf-aig-metadata},
        LiteLLM's \texttt{x-litellm-tags}, and custom
        \texttt{X-Client-Request-Id} headers) already let operators
        annotate requests with environment, team, and workload-type
        tags.  SIRP~\cite{sirp2025} standardizes this with
        classification-axis header fields.
\end{itemize}

\paragraph{Caller identity and authorization scope.}
The four signal classes above describe \emph{what} the request is.
None describes \emph{who} is making it or \emph{what they are
authorized to do}.  The bearer token carried in every LLM API call
fills this gap: in production gateways, it encodes caller identity
\emph{and} authorization scope---which models a key can access, which
tools it may invoke, and what budget it may consume.
\begin{itemize}[nosep]
  \item \textbf{Per-key tool permissions.}
        LiteLLM's Tool Permission Guardrail~\cite{litellm2026toolguard}
        enforces per-key tool allow/deny rules with regex matching on
        tool names and arguments.  Two modes---\emph{block} (reject the
        request) and \emph{rewrite} (strip forbidden tools before the
        model sees them)---demonstrate that caller-scoped tool governance
        is already production-ready at the gateway layer.
  \item \textbf{Structured agent authorization.}
        The Agent Authorization Profile
        (AAP)~\cite{aap2026}---an emerging OAuth~2.0 extension---adds
        five JWT claims: agent identity, capabilities with enforceable
        server-side constraints (domain restrictions, rate limits, time
        windows per tool), task binding (prevents scope drift), delegation
        tracking, and audit trails.  Its core principle is that
        ``authorization must be evaluated at execution time, not just
        during connection.''
  \item \textbf{Infrastructure analogue.}
        Kubernetes RBAC with admission webhooks enforces per-identity
        permissions at the API-server boundary regardless of what the
        application inside the pod believes it should do---the same
        zero-trust principle~\cite{nist800207} applied to LLM
        requests.
\end{itemize}
The bearer token enables a new class of routing decisions absent
from the current WRP roadmap: role-based tool-call enforcement
(\Cref{opp:rbac-enforcement}), per-caller fleet-wide reputation
scoring (\Cref{opp:caller-reputation}), and router-enforced
behavioral commitments (\Cref{opp:behavioral-commitments}).  It
also extends existing opportunities: per-caller cumulative risk
accumulation (\Cref{opp:cumulative-risk}), cross-session token
budgets (\Cref{opp:token-budget}), RBAC predicates in the
governance DSL (\Cref{opp:governance}), role-aware tool catalog
shaping (\Cref{opp:tool-pruning}), and caller-aware memory
isolation (\Cref{opp:gateway-agent-loops}).

\paragraph{Inferred structural signals.}
Beyond explicit metadata, the router can derive workload identity
from request structure at dispatch time:
\begin{itemize}[nosep]
  \item \textbf{Conversation depth.}  The number of message objects
        in the \texttt{messages} array reveals session maturity:
        a 2-message exchange is likely a cold-start chat; a
        20-message array with alternating \texttt{assistant}/\texttt{tool}
        roles is a deep agent session with different model, pool,
        and compression requirements.
  \item \textbf{Tool-call history pattern.}  If prior assistant
        messages contain \texttt{tool\_calls} with high failure rates
        (parseable from subsequent \texttt{tool} role messages
        reporting errors), the router can preemptively route to a
        stronger model or apply compression before the next turn.
  \item \textbf{Reasoning metadata.}  NofT~\cite{noft2026} shows
        that the number of chain-of-thought reasoning steps predicts
        task difficulty pre-inference, enabling difficulty-aware model
        selection with 95\% adversarial detection accuracy.
\end{itemize}

\paragraph{From static to per-stage routing.}
These signals enable a shift from per-request routing (one decision
per API call) to \emph{per-stage} routing within agentic workflows.
Aragog~\cite{aragog2025} demonstrates this: by decoupling a one-time
accuracy-preserving configuration step from a per-stage scheduler
that uses current system observations, it achieves 50--217\%
throughput improvement and 32--79\% latency reduction at peak load.
The WRP framework is well-positioned for this evolution: the
router already maintains session state for multi-turn awareness
(VCD~\cite{vcd2026}, Feedback Detector~\cite{feedbackdet2026}),
and the per-stage identity signals listed above provide the input
features for a stage-aware routing policy that adapts model selection,
compression, and pool assignment as the session evolves.

\subsection{Workload CDF Archetypes}
\label{sec:workload:archetypes}

FleetOpt~\cite{fleetopt2026} identified three prompt-length CDF
archetypes that determine optimal pool and routing configuration:

\begin{enumerate}[nosep]
  \item \textbf{Concentrated-below} (\eg Azure): 80--95\% short
        requests.  Aggressive compression minimizes long-pool cost.
  \item \textbf{Dispersed} (\eg LMSYS): requests span the length
        spectrum.  Balanced two-pool split with moderate compression.
  \item \textbf{Concentrated-above} (\eg SWE-bench): most requests
        are long.  Raise $\boundary_{\text{short}}$ rather than
        compress.
\end{enumerate}

These archetypes are not static: as organizations shift from chat to
agents, the CDF migrates from Archetype~1 toward Archetype~3.
FleetOpt~\cite{fleetopt2026} showed that the minimum-cost fleet is
qualitatively different for each archetype, and
inference-fleet-sim~\cite{fleetsim2026} confirmed that monolithic
topology can outperform disaggregated for Archetype~1 when
$\boundary_{\text{short}} < 4$K.

\section{Dimension 2: The Router Layer}
\label{sec:router}

The router sits between the client and the inference pool.  Our
publications cover three adaptation regimes: static rules, online
bandit adaptation, and RL-based selection.

\subsection{Static Semantic Routing}
\label{sec:router:static}

The \vsr{} position paper~\cite{vllmsr2026} established the
signal-driven architecture: composing thirteen signal types into
deployment-specific policies across vLLM, OpenAI, Anthropic, Azure,
Bedrock, and Gemini backends.  The embedding similarity signal is
powered by mmBERT-Embed~\cite{mmberted2025}, whose 2D Matryoshka
design lets operators dial latency vs.\ quality per deployment
(2.56\,ms at layer-6 for real-time routing, 11\,ms at layer-22 for
maximum recall).  ProbPol~\cite{probpol2026} formalized conflict
detection for these signals.

Token-budget pool routing~\cite{poolrouting2026} implements a
specialized static rule: dispatching to short or long pools based on
estimated total token budget, achieving 17--39\% GPU reduction.
OATS~\cite{oats2026} addresses tool selection with zero-cost
embedding refinement, improving NDCG@5 from 0.869 to 0.940.

Separately, WhenToReason~\cite{wang2025reason} classifies whether a
query warrants reasoning-mode invocation, avoiding unnecessary
chain-of-thought tokens on simple queries.  The Factcheck
Classifier~\cite{factcheck2026} adds another pre-route signal:
queries classified as fact-seeking can be steered toward models with
lower hallucination rates or flagged for post-response validation
via HaluGate~\cite{halugate2025}.

For multi-turn sessions, the \vsr{} project~\cite{vllmsr2026}
provides two mechanisms beyond single-request classification.
First, \emph{context memory injection}: the router accumulates
contextual signals from prior turns (topic, domain, user profile,
difficulty trajectory) and injects this context into the routing
decision for subsequent turns, giving the inference endpoint richer context that improves response
accuracy.  This is similar in spirit to the conversation-state
management in multi-turn APIs like OpenAI's Responses API.  Second, \emph{contrastive embedding classification
for multi-turn safety}: rather than classifying each turn in
isolation, the router uses few-shot contrastive embeddings to detect
jailbreak attempts that spread adversarial content across multiple
turns, which single-turn classifiers cannot detect.

Content safety and privacy form another pre-route gate.  The
MLCommons Safety Level-1 classifier~\cite{safetyl1_2026} performs a
binary safe/unsafe check on every inbound request; flagged requests
are passed to the Level-2 hazard classifier~\cite{safetyl2_2026},
which assigns one of nine MLCommons categories (violent crimes,
non-violent crimes, sex crimes, weapons/CBRNE, self-harm, hate,
specialized advice, privacy, misinformation).  This hierarchical
design keeps latency negligible for safe traffic (only unsafe
requests pay the second-stage cost) and enables per-category routing
policies: privacy-violating prompts can be blocked at the gateway,
misinformation queries routed to fact-grounded models, and
specialized-advice requests logged for compliance auditing.  Both
classifiers use mmBERT LoRA adapters, fitting the project's
low-resource design constraint.

FastRouter~\cite{fastrouter2026} ensures all these routing
decisions---single-turn and multi-turn---happen within the latency
budget: 50\,ms end-to-end with an 800\,MB GPU footprint, small
enough to co-locate with serving.

\subsection{Online Adaptive Routing for Chat}
\label{sec:router:online}

Chat workloads offer a natural feedback signal: the user's next message indicates whether the previous response was
satisfactory.  Our
mmBERT-32K Feedback Detector~\cite{feedbackdet2026} classifies each
user turn into four categories (satisfied, needs clarification, wrong
answer, wants different approach) at 98.8\% accuracy with
$<$1\,ms overhead.  When the detector flags dissatisfaction, the
router can re-route subsequent turns to a stronger model or a
different pool---no explicit reward model is needed because the
user's own reaction serves as the reward.  Combined with the
router's context memory injection (\Cref{sec:router:static}), the
adaptation becomes session-aware: the router knows not just that the
user is dissatisfied, but which topic and difficulty level triggered
the dissatisfaction, enabling targeted re-routing rather than a
blanket model upgrade.

Beyond our project, RouteLLM~\cite{ong2025routellm} learns routers
from human preference data (95\% GPT-4 quality at 26\% cost);
MixLLM~\cite{lu2025mixllm} uses contextual bandits with query tags
(97.25\% of GPT-4 quality at 24.18\% cost);
SageServe~\cite{jia2025sageserve} integrates traffic forecasting
with ILP-based routing (25\% GPU-hour savings at Microsoft scale).

\subsection{RL-Based Routing for Agents}
\label{sec:router:rl}

Agent workloads lack the per-turn user feedback that chat enjoys:
the ``user'' is another LLM or an orchestrator, and task success is
only observable at the end of a multi-step session.  This delayed,
sparse reward makes RL a better fit than bandit methods.
Router-R1~\cite{routerr1_2025} demonstrates the approach: it
instantiates the router itself as an LLM, interleaving ``think''
actions (internal reasoning) with ``route'' actions (model
invocation), and trains end-to-end via PPO with a composite
reward---format, outcome, and cost---that lets the router learn
performance--cost trade-offs without hand-crafted heuristics.
R2-Router~\cite{r2router2026} jointly selects model and output
length budget at 4--5$\times$ lower cost.
M-CMAB~\cite{mcmab2026} uses online RL for multi-modal scheduling
under heterogeneous budgets.

Our AVR system~\cite{avr2026} applies cascading to VLM agent tasks:
Tier~1 multimodal classifier ($\sim$10\,ms), Tier~2 small-VLM
confidence probing with token-filtered scoring, Tier~3 large-model
escalation, projecting up to 78\% cost reduction while staying
within 2\,pp of an all-large-model baseline.  When combined with the
Visual Confused Deputy guardrail~\cite{vcd2026}, high-risk actions
escalate directly to the strongest model.

\paragraph{Adaptation regime depends on workload.}
For \emph{chat}, per-turn user feedback detected by classifiers like
our Feedback Detector~\cite{feedbackdet2026} provides a fast,
low-overhead signal to adjust routing online.  For \emph{agents},
task-level outcomes observed only after multi-step sessions call for
RL-based optimization~\cite{routerr1_2025} that reasons over
sequential routing decisions.

\subsection{Agent Loops, Memory, and Fleet-Scale Orchestration}
\label{sec:router:agent-loops}

Recent agent-serving work argues that inference for tool-using agents
should be treated as a \emph{data-systems} problem, not a sequence of
independent chat completions.  Helium~\cite{helium2026} reframes
agentic LLM serving around workflow structure, reuse, and operator
placement; Sutradhara~\cite{sutradhara2026} co-designs orchestrator
and engine for tool-heavy inference; CONCUR~\cite{concur2026}
applies congestion-aware batching to high-throughput agentic
inference; AgServe~\cite{agserve2025} shows that
\emph{session-aware} scheduling transcends static cost--quality
trade-offs; Continuum~\cite{continuum2025} couples multi-turn
scheduling with KV-cache time-to-live so that paused tool calls do
not silently destroy locality.  These systems align with
experience-driven routing (EvoRoute~\cite{evoroute2026}) and
budget-aware sequential routing (BAAR~\cite{baar2026}), but they are
typically evaluated inside a single stack.

The semantic router complements them by exporting a \emph{fleet-wide}
control surface: the same hooks that implement signal-driven model
routing can (i)~reshape tool and instruction payloads
(ITR-style~\cite{itr2026}, evaluated on that paper's controlled
benchmark), (ii)~attach memory-tier decisions (full
context vs.\ summary vs.\ retrieval-first prefill informed by
MemCost~\cite{memcost2026} and provider prompt-caching studies on
long-horizon agents~\cite{dontbreakcache2026}), and (iii)~emit scheduling hints
(admission deferral, batch affinity) that reduce queueing-induced
retries---a common hidden driver of extra agent rounds.  xRouter's
RL-trained orchestration~\cite{xrouter2025} is evidence that the
router layer can learn cost-aware policies when rewards are defined
over full trajectories; combining that with gateway-native stacks
such as OpenClaw~\cite{openclaw2026}---where real users drive
high-variance multi-turn traffic through a single control
plane---multiplies the diversity of trajectories available for
learning without requiring every framework to share one memory or
tool SDK.

\section{Dimension 3: Pool Architecture}
\label{sec:pool}

Our fleet optimization tools~\cite{fleetopt2026, fleetsim2026,
onewlaw2026} provide the analytical and simulation models for pool
architecture.

\subsection{Homogeneous vs.\ Heterogeneous GPU Pools}
\label{sec:pool:homo-hetero}

FleetOpt~\cite{fleetopt2026} derives the minimum-cost fleet
analytically: a two-pool architecture with optimal boundary
$\boundary^*$ satisfying an equal marginal GPU cost condition.  On
production traces, this reduces GPU cost by 6--82\% versus
homogeneous fleets.  inference-fleet-sim~\cite{fleetsim2026} extends
this to heterogeneous GPU types (A10G, A100, H100) with a
physics-informed performance model.

M\'{e}lange~\cite{griggs2024melange} formulates heterogeneous GPU
allocation as cost-aware bin packing, achieving up to 77\% cost
reduction.  Our simulation tool~\cite{fleetsim2026} shows that the
cheapest GPU type depends on the workload in non-obvious ways:
A10G can beat H100 for concentrated-below workloads where
concurrency matters more than per-token speed.

\subsection{Energy and the 1/W Law}
\label{sec:pool:energy}

The 1/W law~\cite{onewlaw2026} has direct implications for pool
architecture: tokens per watt halves every time the context window
doubles, because of KV-cache concurrency limits.  At 64K
context, an H100 holds 16 sequences ($\tokW \approx 1.5$); at 4K,
256 sequences ($\tokW \approx 17.6$).

Routing topology is a stronger energy lever than GPU generation: two-pool routing gives $\sim$2.5$\times$ better
tok/W than homogeneous; upgrading H100$\to$B200 gives
$\sim$1.7$\times$.  The gains compose multiplicatively to
$\sim$4.25$\times$.  For MoE models like Qwen3-235B-A22B,
active-parameter weight streaming provides a third lever
($\sim$37.8\,tok/W at 8K context, 5.1$\times$ better than dense
Llama-3.1-70B).

Energy optimization therefore requires routing-pool co-design; a
homogeneous fleet cannot reach the energy frontier regardless of GPU
generation.

\subsection{Disaggregated Prefill/Decode and KV-Cache Topology}
\label{sec:pool:disagg}

Splitwise~\cite{patel2024splitwise} and
DistServe~\cite{zhong2024distserve} physically separate prefill and
decode, serving $4.48\times$ more requests at equivalent SLO.
Mooncake~\cite{qin2025mooncake} provides KV-cache-centric
disaggregation with hierarchical offloading (GPU HBM, CPU DRAM, SSD).
NVIDIA Dynamo~\cite{nvidia2026dynamo} enables dynamic scheduling with
KV-cache offloading across memory hierarchies.

Our inference-fleet-sim~\cite{fleetsim2026} models all three
topologies (monolithic, two-pool-routed, disaggregated), revealing
that disaggregation is not always beneficial: for concentrated-below
workloads (Archetype~1) at small $\boundary_{\text{short}}$, the
KV-cache transfer overhead exceeds the benefit.

vLLM's PagedAttention~\cite{kwon2023vllm} reduces KV-cache waste to
below 4\%, and prefix-cache-aware distributed
scheduling~\cite{llmd2026kvcache} delivers $57\times$ speedups.  But
for agentic workloads, standard LRU eviction is
pathological~\cite{vllm2026rfc37003}---context-aware retention
with session-affinity routing is needed.

\section{Cross-Dimensional Interactions: Lessons from Our Work}
\label{sec:cross}

Cross-dimensional interactions are easy to miss when each dimension
is studied separately, but they dominate real-world performance.
Below we formalize the main interactions our publications have
identified.

\subsection{Workload $\times$ Router}
\label{sec:cross:wr}

\begin{observation}[Workload determines router adaptation mechanism]
\label{obs:wr-adapt}
Chat workloads produce per-turn user feedback; our Feedback
Detector~\cite{feedbackdet2026} classifies satisfaction in real
time, allowing the router to re-route on dissatisfaction without
explicit reward modeling.  Agent workloads lack such per-turn
signals---task success is observed only at the end of a multi-step
session---so RL-based routing~\cite{routerr1_2025} that reasons
over sequential decisions is a better fit.  The adaptation
mechanism must match the workload's feedback granularity.
\end{observation}

\begin{observation}[Modality determines routing signals]
\label{obs:wr-modal}
Text-only chat uses embedding similarity and keyword
signals~\cite{vllmsr2026}.  VLM workloads require multimodal
classifiers---SigLIP+MiniLM in AVR~\cite{avr2026}---because
text-only signals cannot assess screenshot complexity.  The router signal set must match the workload modality.
\end{observation}

\begin{observation}[Workload determines routing safety and privacy requirements]
\label{obs:wr-safety}
Different workload types impose different safety and privacy
requirements on the router.  Chat routing errors produce quality
degradation; multi-turn chat adds a subtler threat---adversarial
users can spread jailbreak content across turns, evading
single-turn classifiers.  The \vsr{}~\cite{vllmsr2026} addresses
this with contrastive embedding classification that evaluates turn
sequences, not individual messages.  CUA routing errors produce
outright \emph{security vulnerabilities}: the Visual Confused
Deputy~\cite{vcd2026} shows that misrouting a CUA action to a
less-capable model can cause grounding errors exploitable for
privilege escalation.  Across all workload types, our hierarchical
MLCommons safety pipeline---binary gate~\cite{safetyl1_2026} then
nine-class hazard categorization~\cite{safetyl2_2026}---provides
uniform content filtering at the routing layer, with category-level
granularity that lets operators enforce workload-specific policies
(e.g., blocking privacy-exfiltrating prompts in customer-facing chat,
flagging misinformation in knowledge-retrieval agents).  Safety and privacy should be integrated as routing objectives,
with multi-turn awareness for chat and action-level guardrails for
agents.
\end{observation}

\begin{observation}[Factual workloads require response-time validation]
\label{obs:wr-halu}
Request-time routing alone cannot guarantee correctness: a model may
be well-suited for the query's domain yet still hallucinate on a
specific fact.  The Factcheck Classifier~\cite{factcheck2026}
identifies fact-seeking prompts at request time ($\sim$12\,ms), and
HaluGate~\cite{halugate2025} validates responses at token level
(76--162\,ms overhead).  Together they close a loop that
request-time routing cannot: per-model hallucination rates observed
by HaluGate feed back into the routing policy, gradually steering
factual traffic toward more reliable models.
\end{observation}

\subsection{Router $\times$ Pool}
\label{sec:cross:rp}

\begin{observation}[Router policy co-determines pool sizing]
\label{obs:rp-codesign}
FleetOpt~\cite{fleetopt2026} shows that co-designing the compression
parameter $\gamma$ with pool sizes $(n_s, n_l)$ yields 3.1--6.4\%
lower cost than retrofitting compression onto a pre-existing fleet.
The router is a co-design variable, not a layer atop the pool.
\end{observation}

\begin{observation}[Router latency constrains pool topology]
\label{obs:rp-latency}
FastRouter~\cite{fastrouter2026} showed that routing latency is
a critical constraint: at 4,918\,ms baseline, routing overhead
exceeds the TTFT SLO for short-context requests.  Only after the
98$\times$ reduction to 50\,ms does pool routing become practical
for the full request stream.
mmBERT-Embed~\cite{mmberted2025} reinforces this: its 2D Matryoshka
design lets operators trade embedding quality for latency
(2.56\,ms at layer-6 vs.\ 11\,ms at layer-22), keeping the
similarity signal within the router's latency budget even as pool
topology grows more complex.
\end{observation}

\begin{observation}[Mid-session model switching incurs a KV-cache sunk cost]
\label{obs:rp-sunkcost}
In multi-turn sessions, if the Feedback
Detector~\cite{feedbackdet2026} or context memory signals indicate
that a better model is available, the router faces a dilemma.
Switching models mid-session discards the KV-cache accumulated for
the current model---a \emph{sunk cost} proportional to session
length and context size.  Not switching preserves the cache
investment but forgoes the quality improvement---an \emph{opportunity
cost} that grows with every subsequent turn served by the
suboptimal model.  The optimal policy depends on the remaining
session length, the quality gap between models, and the KV-cache
rebuild cost, creating a Pareto frontier between sunk cost and opportunity cost.
No current system explicitly optimizes this trade-off.
\end{observation}

\subsection{Workload $\times$ Pool}
\label{sec:cross:wp}

\begin{observation}[Workload archetype determines optimal topology]
\label{obs:wp-topo}
Our fleet tools~\cite{fleetopt2026, fleetsim2026} show:
Archetype~1 (concentrated-below) benefits most from two-pool routing.
Archetype~3 (concentrated-above) may benefit from disaggregated
prefill/decode because most requests are long.  Archetype~2
(dispersed) may require multi-tier pooling.
\end{observation}

\begin{observation}[Agent adoption shifts the energy frontier]
\label{obs:wp-energy}
The 1/W law~\cite{onewlaw2026} predicts that shifting from chat
to agents drops fleet-wide energy efficiency by $2$--$4\times$
unless pool topology adapts.  Two-pool routing recovers most of
this loss, but only if pool boundaries track workload evolution.
\end{observation}

\section{Mapping Our Publications to the WRP Matrix}
\label{sec:mapping}

\Cref{tab:interactions} maps each of our publications to the WRP
interaction matrix, showing which cross-dimensional cells each paper
addresses.  The off-diagonal cells that remain sparse represent the
open opportunities in \Cref{sec:vision}.

\begin{table}[t]
\centering
\caption{WRP interaction matrix.  Each cell lists our publications
that address the corresponding cross-dimensional interaction, plus
key external references.  Sparse cells indicate open opportunities.}
\label{tab:interactions}
\small
\renewcommand{\arraystretch}{1.4}
\begin{tabularx}{\textwidth}{>{\bfseries\small}l X X X}
\toprule
& \textbf{Workload} & \textbf{Router} & \textbf{Pool} \\
\midrule
\rowcolor{workloadblue!5}
Workload   & \textit{CDF archetypes}~\cite{fleetopt2026};
            traces~\cite{patel2024splitwise, servegen2025}
  & AVR~\cite{avr2026} (modality);
    VCD~\cite{vcd2026} (safety);
    SafetyL1/L2~\cite{safetyl1_2026, safetyl2_2026} (content safety);
    FeedbackDet~\cite{feedbackdet2026} (chat adapt.);
    HaluGate~\cite{halugate2025} (factual valid.);
    \textit{sparse---see~\Cref{opp:session-cascade,opp:failure-routing}}
  & FleetOpt~\cite{fleetopt2026} (CDF$\to$sizing);
    1/W Law~\cite{onewlaw2026} (energy);
    \textit{sparse---see~\Cref{opp:proactive-scaling}} \\
\rowcolor{routergreen!5}
Router & FleetOpt~\cite{fleetopt2026} (compression changes CDF);
         WhenToReason~\cite{wang2025reason}
  & \vsr{}~\cite{vllmsr2026} (signals);
    ProbPol~\cite{probpol2026} (conflicts);
    FastRouter~\cite{fastrouter2026} (latency);
    mmBERT-Embed~\cite{mmberted2025} (embeddings);
    OATS~\cite{oats2026} (tools);
    SemCache~\cite{wang2025cache} (caching);
    HaluGate~\cite{halugate2025} (validation)
  & FleetOpt~\cite{fleetopt2026} (co-design $\gamma$);
    \textit{sparse---see~\Cref{opp:dynamic-boundary,opp:kv-retention}} \\
\rowcolor{poolorange!5}
Pool & FleetSim~\cite{fleetsim2026} (topology$\to$cost);
       1/W~\cite{onewlaw2026} (topology$\to$energy)
  & \textit{sparse---see~\Cref{opp:pool-feedback}}
  & FleetSim~\cite{fleetsim2026} (A10G/A100/H100);
    M\'{e}lange~\cite{griggs2024melange};
    DistServe~\cite{zhong2024distserve} \\
\bottomrule
\end{tabularx}
\end{table}

\section{Vision: A Research Roadmap}
\label{sec:vision}

The WRP matrix in \Cref{tab:interactions} reveals two structurally
sparse cells---Router $\times$ Pool and Pool $\times$ Router---and
several interactions in the dense cells that our publications have
identified but not yet exploited.  Below we propose twenty-one research
directions that follow directly from these gaps, each framed around
what the semantic router's signal architecture and the WRP
cross-dimensional view uniquely enable.

\Cref{tab:maturity} assigns each opportunity a maturity tier.
Throughout this section, projected benefits are estimates grounded
in our prior measurements unless a specific external citation is
given.

\begin{table}[t]
\centering
\caption{Maturity and feasibility of the twenty-one proposed
opportunities.  \textbf{Tier} indicates the primary barrier:
\emph{engineering} (building blocks exist, integration needed),
\emph{research} (open technical questions).}
\label{tab:maturity}
\small
\renewcommand{\arraystretch}{1.15}
\begin{tabularx}{\textwidth}{c l c X}
\toprule
\textbf{\#} & \textbf{Opportunity} & \textbf{Tier} & \textbf{Primary barrier} \\
\midrule
\rowcolor{routergreen!8}
2  & HaluGate model reputation routing    & Engineering & Sample size per (model, domain) cell \\
5  & Token-budget enforcement for agents   & Engineering & Domain-specific budget model calibration \\
\rowcolor{routergreen!8}
8  & Follow-the-sun pool rebalancing       & Engineering & Sliding-window CDF estimation + FleetOpt \\
12 & KV-cache retention directives         & Engineering & Retention-directive API integration \\
\rowcolor{routergreen!8}
15 & Governance-as-code                    & Engineering & Extend DSL with WRP constraints \\
19 & Request-level RBAC enforcement         & Engineering & Gateway hook + JWT integration \\
\midrule
\rowcolor{poolorange!8}
1  & Offline RL for session-length routing  & Research & Credit assignment + off-policy evaluation \\
3  & Cumulative multi-turn risk scoring     & Research & Threshold calibration on adversarial benchmarks \\
\rowcolor{poolorange!8}
4  & Failure-class-aware routing            & Research & Multi-dimensional failure weighting \\
\textbf{17} & \textbf{Fleet-learned joint tool--model optimization} & \textbf{Research} & \textbf{Joint (tool, model) outcome table sparsity} \\
\rowcolor{poolorange!8}
18 & Gateway-coordinated agent loops       & Research & End-to-end fleet integration unevaluated; subsystems validated separately \\
\rowcolor{poolorange!8}
20 & Fleet-wide caller reputation           & Research & Heterogeneous security-event scoring + decay calibration \\
21 & Router-enforced behavioral commitments & Research & Boundary-declaration NLU classifier + false-positive risk \\
\rowcolor{poolorange!8}
6  & Pool-state inference from observables  & Research & Prefill-rate regression + cold start \\
7  & Output-length-aware pool routing       & Research & Output-length prediction variance \\
\rowcolor{poolorange!8}
9  & Pool-aware cascading                   & Research & Cascade-abort threshold + grace period \\
10 & Online $(\gamma, \boundary^*)$ co-adaptation & Research & Oscillation control + validation gap \\
\rowcolor{poolorange!8}
11 & Mixed-archetype fleet provisioning     & Research & Multi-tenant FleetOpt extension \\
13 & Archetype-driven proactive scaling     & Research & Archetype-aware forecast vs.\ aggregate baseline \\
\rowcolor{poolorange!8}
14 & Energy as a routing objective          & Research & tok/W signal calibration per pool config \\
16 & Closed-loop self-adaptation            & Research & Knob interaction strength + composite reward \\
\bottomrule
\end{tabularx}
\end{table}

\subsection{Workload $\times$ Router}

\begin{opportunity}[Offline RL for session-length-minimizing routing]
\label{opp:session-cascade}
In agent workloads, a long session is often the consequence of poor
routing decisions at early turns.  If the router selects a weak
model at turn~2 and the model fails a tool call, the agent retries
at turn~3 with a longer prompt, which fails again, and so on.  A
session that could have completed in 3~turns with the right model
and tool selection at the outset may instead take 8~turns---each
additional turn growing the prompt
(ServeGen~\cite{servegen2025} reports 3--10$\times$ more LLM calls
per user request for agents than chatbots) and consuming
proportionally more compute.

The router observes the full trajectory of every completed session:
which model served each turn, which tools were invoked
(OATS~\cite{oats2026}), whether each tool call succeeded, and how
many turns the session took.  This is a natural offline RL dataset.
The state is (turn number, tools tried so far, cumulative context
length, domain); the action is (model, tool) selection; the reward
is negative session length (or task-completion / session-length for
quality-adjusted optimization).  The workload identity signals from
\Cref{sec:workload:identity} enrich this state space at zero
inference cost: the system-prompt fingerprint provides an instant
domain prior, the tool-definition catalog identifies the agent's
operational mode, and the conversation depth from the messages
array gives exact session maturity---all available before the
router inspects content.  The router's fleet-wide visibility
is the structural advantage over agent-side SDKs: it collects
trajectories from thousands of sessions across all tenants, giving
it the training volume that no single agent instance can match.

An offline RL policy (e.g., conservative Q-learning or decision
transformer) trained on these trajectories learns which early-turn
decisions lead to shorter sessions: ``for coding tasks, routing
turn~1 to the frontier model and selecting \texttt{semantic\_search}
over \texttt{grep} reduces median session length from 7 to
4~turns.''  The learned policy replaces OATS's per-turn greedy
selection with a \emph{per-stage} strategy that adapts model and
tool selection as the session evolves---the same progression
Aragog~\cite{aragog2025} demonstrates for throughput (50--217\%
improvement by decoupling one-time configuration from per-stage
scheduling), applied here to session-length minimization.

\paragraph{Intervention actions: cut-and-regenerate.}
The action space above covers \emph{forward} decisions---which model
and tools for the \emph{next} turn.
Search-R2~\cite{searchr2_2026} demonstrates that adding a
\emph{corrective} action---``cut the trajectory at turn~$k$ and
regenerate the suffix''---yields consistent 5--5.5\,pp accuracy gains
over forward-only policies on search-integrated reasoning benchmarks.
The router's offline RL dataset already contains the signal for
learning this action: session logs record per-turn outcomes (model,
tool, success/failure, token cost), enabling hindsight labeling of
the \emph{regeneration gain} $G_k$---the difference between the
actual cost from turn~$k$ onward and the expected cost of
regenerating from $k$ with the fleet's best model for that
(domain, turn-depth) cell.  Adding \texttt{intervene($k$, model)}
to the action space lets the RL policy learn not only ``which model
for the next turn'' but also ``should I cut back to a previous
turn and regenerate?''  The handoff-penalty matrix
(\Cref{opp:failure-routing}) provides the cost of intervention;
the policy learns to intervene only when $G_k$ exceeds the handoff
penalty plus recompute cost.

\begin{frisk}
The router already logs per-turn model and tool-call outcomes for
OATS and the Feedback Detector; assembling these into session
trajectories is bookkeeping.  Offline RL is well suited to this
setting: recent work shows that conservative Q-learning variants
learn effective scheduling policies from logged data
alone~\cite{offlinerl2025scheduling}, generalizing even from
suboptimal (random-heuristic) training trajectories---closely
matching the quality of router logs collected under a non-optimal
greedy policy.  The research questions are:
(a)~credit assignment---which of the 8~turns caused the extra
5~turns?  Offline RL handles this via temporal-difference learning,
but noisy agent environments may require variance reduction.
(b)~Off-policy evaluation---the learned policy must be validated
before deployment.  Importance-weighted estimators on held-out
sessions provide offline evaluation without live traffic risk.
The comparison baseline is OATS with single-turn greedy selection;
the metric is median session length (turns and tokens) for
completed tasks on SWE-bench and BFCL.
\end{frisk}
\end{opportunity}

\begin{opportunity}[HaluGate-driven model reputation routing]
\label{opp:halugate-reputation}
The Factcheck Classifier~\cite{factcheck2026} identifies
fact-seeking prompts at request time ($\sim$12\,ms).
HaluGate~\cite{halugate2025} validates responses at token level
(76--162\,ms overhead) and produces a per-response hallucination
verdict.  Today, these verdicts are logged but not fed back into
routing decisions.

The router processes every response through HaluGate for
fact-seeking traffic.  Over time, it accumulates (model, domain,
hallucinated) triples---enough to maintain a per-model,
per-domain accuracy estimate.  For example: ``Model~A hallucinates
on medical queries 12\% of the time but only 2\% on coding;
Model~B is the reverse.''  The opportunity is to use this estimate
as a live routing signal: for each fact-seeking request, the
router queries the per-model accuracy table for the request's
domain and routes to the model with the lowest hallucination rate,
subject to cost and latency constraints.

\begin{frisk}
The data collection is zero-overhead (HaluGate already runs).
The per-model accuracy table is a counter per (model, domain) pair,
updated on each verdict.  The engineering question is sample
size: how many verdicts per (model, domain) cell are needed before
the estimate is reliable?  A Bayesian approach (Beta prior,
Bernoulli likelihood) gives calibrated confidence intervals from
small samples.  The risk is distributional shift: model updates
invalidate historical accuracy estimates.  A sliding window
(e.g., last 1,000 verdicts per cell) handles this.
\end{frisk}
\end{opportunity}

\begin{opportunity}[Cumulative multi-turn risk scoring]
\label{opp:cumulative-risk}
SafetyL1~\cite{safetyl1_2026} and SafetyL2~\cite{safetyl2_2026}
classify each turn independently.  Multi-turn jailbreak
attacks~\cite{vcd2026} spread malicious intent across turns so that
no single turn triggers the classifier---each turn is individually
benign, but the cumulative intent is harmful.  The
\vsr{}~\cite{vllmsr2026} addresses this partially with contrastive
embedding classification that evaluates turn sequences, but the
routing response to detected escalation is not formalized.

The opportunity is to maintain a per-session cumulative risk score
that aggregates borderline SafetyL1/L2 signals across turns.
SafetyL1 produces a continuous confidence score (not just a binary
verdict); a turn that scores 0.4 (below the 0.5 threshold) is
not flagged, but three consecutive 0.4-scoring turns represent
genuine escalation.  The cumulative score is the EMA of per-turn
SafetyL1 confidences.  When it crosses a configurable threshold,
the router takes graduated action: route to a safety-specialized
model, reduce the model's sampling temperature via the API, or
flag the session for human review.

\paragraph{Per-caller accumulation across the fleet.}
The bearer token (\Cref{sec:workload:identity}) extends this
per-session mechanism to a \emph{per-caller} dimension.  A non-owner
whose interactions trigger borderline SafetyL1 scores across five
different agents accumulates a high fleet-wide risk score even if no
single session crosses the threshold---the same Crescendo-style
escalation applied across agents rather than across turns.  The
mechanism is identical (EMA), keyed on \texttt{(bearer\_token)} in
addition to \texttt{(session\_id)}.  Real-world multi-agent attacks
validate this threat model: in live deployments the same non-owner
attacked multiple agents sequentially, each encountering the attack
fresh with no shared threat memory~\cite{agentschaos2026}.

\paragraph{Context-surgical intervention via cut-and-regenerate.}
The graduated response above---escalate model, reduce temperature,
flag session---routes the \emph{next} turn to a safer model but
leaves adversarial content from prior turns in the conversation
history.  A sophisticated multi-turn attacker (Crescendo-style) has
already planted adversarial framing across several turns; routing
turn~6 to a safety model with turns 2--5 in context may not help
because the model still processes the adversarial setup.
Search-R2~\cite{searchr2_2026} formalizes an alternative:
instead of continuing with a polluted context, the Meta-Refiner
\emph{cuts} the trajectory at the seed turn (the earliest turn where
SafetyL1 confidence dropped below threshold) and regenerates from the
clean prefix.  The adversarial content is removed entirely, not merely
overridden by a stronger model.  This maps to the content integrity
levels in \Cref{opp:gateway-agent-loops}: turns that triggered safety
escalation are classified as Tier~0 (Hostile) and excluded from the
forwarded context---the ``no write up'' rule applied to conversation
history.  The enforcement mechanism is the same
\texttt{async\_pre\_call\_hook} used for tool catalog shaping: the
router strips flagged turns from the messages array before forwarding
to the new model.

\begin{frisk}
SafetyL1 already produces continuous confidence scores; computing
the EMA is trivial.  The threat model is grounded in Crescendo
attacks~\cite{crescendo2024}, which achieve 29--61\% higher
jailbreak rates than single-turn methods by gradually escalating
across benign-looking turns.  Recent proxy-level defenses validate
the cumulative-scoring approach: a peak$+$accumulation formula
combining single-turn peak risk, persistence ratio, and category
diversity achieves 90.8\% recall at 1.20\% false-positive rate on
10,654 multi-turn conversations~\cite{peakaccum2026}; DeepContext
uses recurrent networks to track temporal intent drift with sub-20\,ms
overhead and 0.84~F1~\cite{deepcontext2026}.  The cut-and-regenerate
mechanism is grounded in Search-R2~\cite{searchr2_2026}, which proves
that surgical intervention outperforms trajectory-level rejection on
search-integrated reasoning tasks; the safety analog replaces the
``is the answer correct?'' discriminator with the cumulative risk
score---structurally identical.  The research question
is threshold calibration: too aggressive triggers false positives on
legitimate multi-turn conversations that happen to touch sensitive
topics; too lenient misses real attacks.  Validation on multi-turn
adversarial benchmarks (e.g., the 7,152-prompt MultiBreak dataset)
is the natural evaluation path.  The graduated response
(escalate model $\to$ reduce temperature $\to$ cut-and-regenerate
$\to$ flag session) provides defense-in-depth rather than a binary
block.  The per-caller
extension adds a fleet-wide dimension: the calibration question
becomes whether the cross-agent EMA decay rate should differ from
the within-session decay rate (likely yes---cross-agent escalation
is slower but more deliberate).
\end{frisk}
\end{opportunity}

\begin{opportunity}[Failure-class-aware routing from router outcome memory]
\label{opp:failure-routing}
ErrorAtlas~\cite{erroratlas2026} shows that each LLM has a unique
``failure signature''---recurring error patterns that vary by model
and domain---across 83~models and 35~datasets.  An oracle selector
that always picks the best model outperforms the single best model,
yet production routers capture only $\sim$89\% of oracle
utility~\cite{llmrouterbench2026}.  The gap comes from failure modes
the router does not track: wrong tool selection (35\% of tool-call
failures~\cite{agenticfaults2026}), incorrect parameters
(25--68\%), retry loops where weaker models degenerate into
repetitive re-invocations (baseline recovery rate 32.76\% vs.\ 89.68\%
with targeted intervention~\cite{paladin2025}), and mid-session
model-switch degradation ($\pm$4--13\,pp performance swing when
the suffix model continues a foreign prefix~\cite{handoffdrift2026}).

The vSR already observes the outcome of every request through
HaluGate (hallucination verdict), SafetyL1/L2 (safety score),
OATS (tool-call success/failure), and the Feedback Detector (quality
signal).  The workload identity signals from
\Cref{sec:workload:identity} add a pre-inference dimension: the
\texttt{tools} array in the request reveals the agent's available
tool set, and the tool-call history in prior \texttt{tool}-role
messages exposes failures from earlier turns \emph{before} the
router selects the next model.  Over thousands of requests, these
produce a per-\emph{(model, domain, tool-set, failure class)}
outcome table:
\begin{itemize}[nosep]
  \item Model~A, coding, with \texttt{[grep, semantic\_search,
        bash]}: hallucination 2\%, tool-call success 91\%,
        retry-loop rate 8\%;
  \item Model~B, medical, with \texttt{[search, calculator]}:
        hallucination 3\%, tool-call success 85\%,
        retry-loop rate 4\%;
  \item Model~C, coding, with \texttt{[semantic\_search]} only:
        tool-call success 94\%.
\end{itemize}
The tool-set dimension is critical: the same model may excel with
one tool combination and fail with another, and this information is
available at zero cost from the request structure.
The routing decision becomes: for this (domain, task-type), which
model minimizes \emph{combined} failure risk?  This generalizes
\Cref{opp:halugate-reputation} from hallucination alone to a
multi-dimensional failure profile.

A second dimension is \emph{handoff penalty awareness}.  The router
tracks what happens when it switches models mid-session: (Model~A
$\to$ Model~B) in coding sessions adds $+2.1$ mean turns, while
(Model~A $\to$ Model~C) adds only $+0.3$ turns.  It avoids costly
switches: if Model~A served turns 1--3 and Model~B has a known
handoff penalty for Model~A's prefix, the router stays with Model~A
or selects a low-penalty alternative.

A third dimension is \emph{silent drift detection}.  The same
per-(model, domain) statistics, monitored as a time series, detect
quality regressions from provider-side model updates---a real
production concern (documented March 2026 incident where a frontier
model silently regressed to mid-tier quality without code changes).
A 3$\sigma$ deviation from the sliding-window baseline triggers
automatic traffic rebalancing.

A fourth dimension is \emph{cut-and-regenerate intervention}.  When
HaluGate or OATS flags a failure mid-session, the current response is
to retry the turn or switch models going forward---both approaches
leave the failed content in the conversation history, where it causes
self-conditioning errors on subsequent
turns~\cite{driftbench2026}.  Search-R2~\cite{searchr2_2026}
formalizes a better strategy for search-integrated reasoning:
decompose the process into an Actor (generates trajectories) and a
Meta-Refiner that diagnoses and repairs flawed steps via a
``cut-and-regenerate'' mechanism---preserving the valid prefix while
surgically replacing the flawed suffix.  Across seven QA benchmarks
and three model scales (7B--32B), this produces consistent
5--5.5\,pp accuracy gains over the base Actor with minimal overhead.

The semantic router is structurally positioned as the fleet-wide
Meta-Refiner: it already runs the \textbf{Discriminator}
(HaluGate, SafetyL1/L2, OATS, Feedback Detector detect that a
trajectory has deviated) and has the data for the \textbf{Trimmer}
(the failure-class table identifies which turn is the likely root
cause).  The missing piece is the \textbf{cut-and-regenerate action}:
when the Discriminator flags turn~$k$, the router strips turns
$k$--$T$ from the conversation history, selects the best-fit model
from the per-(domain, failure-class) reputation table, and re-routes
turn~$k$ with a clean prefix.  The expected regeneration gain
$G_k = \text{cost}(k \to T\text{, actual}) -
\text{cost}(k \to T\text{, counterfactual})$ is estimable from
fleet-wide session logs; the router intervenes only when $G_k$
exceeds the handoff penalty.

The router's structural advantage over agent-side SDKs is fleet-wide
visibility: it builds the (model, domain, failure-class) table from
all tenants' sessions, a volume no individual agent can match.
Research validates this asymmetry: agents confined to single-session
reflection produce narrow corrections that often repeat the same
mistakes~\cite{samule2025}, while cross-session error aggregation
yields transferable failure schemata that improve error localization
by up to 19.8\%~\cite{correct2025}.  The router provides this
cross-session intelligence as a shared service---new agents benefit
immediately from fleet-wide failure statistics rather than facing a
cold-start learning problem.

\begin{frisk}
The per-(model, domain, failure-class) counters are zero-overhead
(all signals are already computed).  The handoff-penalty matrix
requires multi-turn session correlation, which the router already
performs for VCD and the Feedback Detector.
The cut-and-regenerate mechanism is grounded in
Search-R2~\cite{searchr2_2026}, which proves that the expected reward
gain decomposes as $\Delta J = A_{\mathrm{prec}} +
V_{\mathrm{inter}} \times S_{\mathrm{trim}}$, where
$A_{\mathrm{prec}}$ is the discriminator's precision (positive if the
discriminator reliably separates good from bad trajectories),
$S_{\mathrm{trim}}$ is the trimmer's skill (positive if it cuts at the
step with highest regeneration gain), and $V_{\mathrm{inter}}$ is the
intervention volume.  All three conditions are satisfiable at the
router layer: the discriminator (HaluGate/OATS) is already validated,
the trimmer leverages fleet-wide failure-class statistics, and the
intervention threshold is self-calibrating via the handoff-penalty
matrix.
The research questions
are: (a)~how many observations per cell before the failure-class
estimate is reliable (Bayesian Beta-Bernoulli gives calibrated
intervals from small samples), (b)~how to weight multiple failure
dimensions into a single routing score (Pareto-front analysis or
a learned weighting from session-outcome data), (c)~how
quickly the drift detector must react without over-triggering on
natural variance, and (d)~whether the regeneration gain $G_k$
computed from fleet logs is a reliable predictor of actual
improvement for a specific session (the counterfactual is
unobserved---importance-weighted estimators from the offline RL
literature, \Cref{opp:session-cascade}, provide the evaluation
methodology).  The baseline is single-dimensional HaluGate
reputation routing (\Cref{opp:halugate-reputation}); the metric
is oracle-utility capture rate on LLMRouterBench.
\end{frisk}
\end{opportunity}

\begin{opportunity}[Runtime token-budget enforcement for agent sessions]
\label{opp:token-budget}
Agent workflows face a token-budget explosion: a single user request
triggers 3--10$\times$ more LLM calls than a chatbot
request~\cite{servegen2025}, with context growing quadratically as
each turn re-sends the full message history.  A 10-step agent with a
4K-token system prompt and 500-token tool outputs consumes over
40K~input tokens from context carriage alone.  Infinite retry loops
(failure class~F4 in the taxonomy
of~\cite{agenticfaults2026}) compound this: an unconstrained
agent can consume 50$\times$ the tokens of a single linear pass.

The root cause is often upstream: hallucinations, wrong tool
selections, and tool-call parameter errors inject useless content
into the conversation history that is re-sent on every subsequent
turn.  Failed attempts consume 20--40\% of total tokens with zero
productive output, and context-window utilization falls below 30\%
useful content after accounting for repeated system prompts and stale
error traces.  Localizing the faulty step is itself unreliable---even
frontier models achieve only 41.1\% accuracy at hallucination
attribution, dropping to 11.6\% for tool-use
errors~\cite{agenthallu2026}---so agents resort to expensive
full-context re-summarization.  The cost compounds through
self-conditioning: an LLM processing its own prior errors becomes
measurably more likely to produce further
errors~\cite{driftbench2026}, creating a vicious cycle where each
failed turn both wastes tokens and increases the probability of the
next failure.

Current mitigations live in the agent SDK (per-workflow budget caps)
or a separate billing layer (per-tenant daily limits).  Neither has
access to the router's session trajectory data: turn number,
cumulative tokens dispatched, tool-call outcomes per turn, and the
domain-specific session-length distribution learned from historical
sessions.  Agent-side compression frameworks such as
ACON~\cite{acon2025} (26--54\% peak-token reduction) and
Focus~\cite{focus2026} (22.7\% savings on SWE-bench) show that
compression is effective, but they optimize a single session in
isolation---the router's fleet-wide visibility enables
\emph{domain-specific} compression policies learned from outcome
data across thousands of sessions.

The router can enforce token budgets at the dispatch layer.  For each
active session, it maintains a running token counter (prompt $+$
completion tokens, updated on each dispatch and completion event).
The system-prompt fingerprint (\Cref{sec:workload:identity})
provides instant budget-model selection: a hash lookup maps each
deployment to its historical session-cost distribution, skipping
domain classification entirely.  The conversation depth from the
messages array gives exact session maturity---the router knows
it is at turn~6 of a session whose P90 cost at turn~6 is $X$~tokens.

When a session exceeds 3$\times$ the expected budget at its
current turn, the router applies a graduated response:
\begin{enumerate}[nosep]
  \item \textbf{Shape tools}: reduce the \texttt{tools} array to
        the highest-value subset for this domain and turn, based
        on fleet-wide tool outcome data
        (\Cref{opp:tool-pruning})---ITR~\cite{itr2026} shows this
        alone recovers up to 95\% of per-step context tokens while
        SMART~\cite{smart2026} shows 37\% performance gain;
  \item \textbf{Compress}: apply FleetOpt's $\gamma$ to reduce the
        remaining prompt length;
  \item \textbf{Downgrade}: route the next turn to a cheaper model
        (the session has already exceeded its cost budget; frontier
        compute is not justified);
  \item \textbf{Terminate}: if the above steps fail to
        arrest growth, return a budget-exceeded signal to the
        orchestrator.
\end{enumerate}
The router is the natural enforcement point because it sees the
actual token flow across all turns, while the agent SDK sees only its
own session and the billing layer operates on aggregated cost with
minutes of delay.

\paragraph{Model-aware compression.}
The graduated response above compresses and downgrades without
considering which model will serve the compressed request.  But models
differ dramatically in their effective context
capacity---RULER~\cite{ruler2024} evaluated 17~models and found that
despite all claiming 32K+~token contexts, only half maintained
satisfactory performance at 32K; effective lengths typically reach
50--65\% of the advertised window, with sudden rather than gradual
degradation.  L2A~\cite{l2a2026} shows that modern attention
mechanisms can skip global attention for $\sim$80\% of tokens with
negligible accuracy loss, but which 80\% depends on the architecture
(MoE vs.\ dense, GQA vs.\ MHA, sliding-window vs.\ full attention).

This means the compress$\to$downgrade cascade should be
\emph{model-conditioned}: a 128K-effective model (e.g.,
Jamba-1.5) can handle the full agent history with light compression,
while a 32K-effective model needs aggressive compression of the same
history---and for certain tasks, the aggressively compressed input to
the cheaper model may be \emph{more cost-effective} than the full
input to the expensive model.
AgentCompress~\cite{agentcompress2026} validates this coupling: a
lightweight neural controller that estimates task complexity and
routes to an appropriately compressed model version reduces cost by
68.3\% while preserving 96.2\% of the original success rate.  The
router extends this from per-model quantization levels to the full
WRP decision: (compression level, target model, target pool) selected
jointly from the fleet-wide per-\emph{(domain, context\_length,
model)} outcome table.

\paragraph{Per-caller cross-session budgets.}
The bearer token (\Cref{sec:workload:identity}) extends per-session
enforcement to \emph{per-caller cross-session} budgets.  The
mechanism is direct: \texttt{sum(tokens consumed by bearer\_token\_X
across all agents) > non\_owner\_budget\_ceiling}.  Live agent
deployments confirm the need: one non-owner consumed 60K~tokens over
9~days across multiple sessions, and another uploaded 10\,MB of
attachments in a single burst~\cite{agentschaos2026}.  LiteLLM
already supports per-key monetary budget limits in production; the
extension from monetary to token/tool-invocation budgets with
role-dependent ceilings follows the same infrastructure, keyed on the
bearer token's identity claims.

\begin{frisk}
The per-session token counter is zero-overhead (the router already
tracks dispatches and completions).  The graduated
compress$\to$downgrade$\to$terminate response mirrors the
congestion-control paradigm that is now standard for LLM rate
limiting: token-bucket algorithms enforce per-tenant budgets in
production gateways, and Concur~\cite{concur2026} applies
congestion-control-inspired feedback to regulate agent concurrency
at the engine level.  The domain-specific budget model is a
per-(domain, turn-number) percentile table, updated from the same
session data used for offline RL (\Cref{opp:session-cascade}).
The risk is false positives: a legitimately long session (e.g., a
complex SWE-bench task) may be terminated prematurely.  Using P90
rather than P50 as the baseline and applying the graduated
response provides a safety margin.  Fact-based memory
systems~\cite{memcost2026} offer a complementary cost lever: at
100K-token context lengths, they become cheaper than long-context
inference after $\sim$10 turns, precisely the regime where agent
sessions operate.  The router can trigger a memory-system handoff
when context carriage cost exceeds the memory-system break-even
point.  The comparison baseline is SDK-side per-workflow budget
caps; the metric is token waste reduction (tokens consumed by
sessions that ultimately fail) on SWE-bench and BFCL traces.
\end{frisk}
\end{opportunity}

\begin{opportunity}[Fleet-learned joint tool--model selection optimization]
\label{opp:tool-pruning}
Agent frameworks send the full tool catalog on every LLM call,
rely on the LLM to select the right tool, and resolve every tool
call with the same model.  This is triply wasteful: the monolithic
catalog consumes up to 90\% of context tokens~\cite{itr2026};
per-turn greedy tool selection ignores the aggregate cost--accuracy
outcome of the full tool-call \emph{sequence}; and using a single
model for all tool calls ignores the 250$\times$ cost range across
model tiers and the fact that different tools demand different model
capabilities (a \texttt{web\_search} summary needs only a cheap
text model; an \texttt{image\_analysis} call requires a vision
model; a \texttt{code\_interpreter} invocation benefits from a
reasoning model).  A session that uses tool~A resolved by model~X
at turn~3 may complete in 5~turns; using tool~B with model~Y at the
same turn may cause a failure that triggers 3~retries and doubles
the session cost.  The goal is not necessarily fewer tools or fewer
turns---it is the \emph{(tool, model)} strategy that minimizes
aggregate session cost while maximizing task completion.

Computer-use agents illustrate this coupling vividly.  The same
web task---e.g., ``find the cheapest flight on this page''---can be
solved via two (tool, model) paths: (a)~a \emph{vision} path that
screenshots the page and sends it to a multimodal model
($\sim$\$0.001/page in vision tokens, but requires an expensive VLM
for grounding), or (b)~a \emph{DOM/text} path that extracts the
accessibility tree or HTML and processes it with a cheap text model
($\sim$\$0.15/page in text tokens, but no vision model needed).
AVR~\cite{avr2026} demonstrates this exact routing problem:
adaptive model selection per GUI action reduces inference cost by up
to 78\% while staying within 2\,pp of an all-large-model baseline.
A systematic comparison of web agent interfaces~\cite{webinterface2025}
quantifies the gap: HTML-browsing agents consume 241K~tokens per
task at F1~0.67, while structured-interface agents achieve
F1~0.75--0.87 at 47K--140K~tokens.  The optimal (tool, model) path
depends on the page, the task, and the agent's history---exactly the
kind of context that fleet-wide data can disambiguate.

Two independent research threads validate that both halves of this
joint action space are learnable from historical data:

\paragraph{Tool selection is learnable.}
\begin{enumerate}[nosep]
  \item \textbf{Tool transition patterns are predictable.}
        AutoTool~\cite{autotool2025} constructs directed graphs from
        historical agent trajectories and finds that tool invocations
        follow predictable sequential patterns (``tool usage
        inertia''), reducing inference cost by 30\%.
  \item \textbf{Recurring sequences can be bundled.}
        AWO~\cite{awo2026} discovers that 14.3\% of tasks follow
        identical execution trajectories after 5~steps; transforming
        recurring sequences into deterministic meta-tools reduces LLM
        calls by 11.9\% while increasing task success by 4.2\,pp.
  \item \textbf{Fine-grained tool rewards produce large gains.}
        ToolRLA~\cite{toolrla2026} decomposes the reward into tool
        selection $\times$ parameter accuracy $\times$ compliance;
        deployed in production (80+ advisors, 1,200+~daily queries),
        it achieves 47\% task-completion improvement and 63\%
        tool-error reduction.
  \item \textbf{Cost awareness changes the optimal strategy.}
        BATS~\cite{bats2025} shows that budget-unaware agents hit
        performance ceilings regardless of tool budget; a budget
        tracker that signals remaining resources enables dynamic
        strategy adaptation---``dig deeper'' vs.\ ``pivot''---pushing
        the cost--performance Pareto frontier.
\end{enumerate}

\paragraph{Per-step model selection is learnable.}
\begin{enumerate}[nosep]
  \item \textbf{Per-step Pareto-optimal model selection.}
        EvoRoute~\cite{evoroute2026} dynamically selects model
        backbones at each agent step using experience-driven
        retrieval and Thompson sampling, reducing execution cost by
        up to 80\% and latency by over 70\% on GAIA and BrowseComp+
        benchmarks while sustaining task performance.
  \item \textbf{Budget-aware sequential model routing.}
        BAAR~\cite{baar2026} formalizes agentic model routing as a
        sequential, path-dependent decision problem: early routing
        mistakes compound, and feedback arrives only at task end.
        Boundary-guided policy optimization improves the cost--success
        frontier under strict per-task budgets (Microsoft).
  \item \textbf{Specialized model pipeline assembly.}
        HierRouter~\cite{hierouter2025} trains a PPO-based RL agent
        to iteratively select which specialized model to invoke at each
        stage of multi-hop inference, conditioning on cumulative cost
        and evolving context.  It improves response quality by up to
        2.4$\times$ compared to single-model baselines at minimal
        additional cost.
\end{enumerate}

\noindent
All seven approaches operate on the agent side, learning from a
single agent's or benchmark's trajectories.  The router's structural
advantage is fleet-wide visibility: it observes (tool, model)
outcomes across all tenants and accumulates a
per-\emph{(domain, turn-number, tool, model)} outcome table---tool
success rate, model accuracy, downstream retries, and total cost
contribution---from thousands of sessions.  AutoTool's transition
graph, learned per-agent, becomes a fleet-wide transition graph
trained on orders of magnitude more data; EvoRoute's Pareto
filtering, performed per-benchmark, operates over production-scale
outcome tables.

The router acts on this knowledge through four mechanisms at
dispatch time:
\begin{itemize}[nosep]
  \item \textbf{Role-aware tool filtering.}
        Before any efficiency-driven optimization, the bearer
        token's authorization scope
        (\Cref{sec:workload:identity}) restricts the tool catalog
        to tools the caller is permitted to invoke.  Non-owner
        requests get a \emph{restricted} catalog (only safe,
        read-only tools), not just an \emph{optimized} one.
        LiteLLM's ``rewrite'' mode~\cite{litellm2026toolguard}
        already implements this at the gateway: unauthorized tools
        are stripped from the \texttt{tools} array before the model
        sees them, using the same
        \texttt{async\_pre\_call\_hook}
        that powers efficiency-driven shaping.
        Authorization-driven filtering applies \emph{before}
        efficiency-driven filtering, ensuring that safety constraints
        are never relaxed by cost optimization.
  \item \textbf{Tool catalog shaping.}  Expose the top-$k$ tools
        ranked by fleet-wide success rate for the current (domain,
        turn) cell, reducing context overhead (ITR~\cite{itr2026}
        validates 95\% token reduction agent-side;
        SMART~\cite{smart2026} shows 37\% performance gain from
        24\% tool-use reduction).
  \item \textbf{Tool-conditional model routing.}  Select the
        resolving model based on the tool: a vision model for
        \texttt{image\_analysis}, a reasoning model for
        \texttt{code\_interpreter}, a cheap text model for
        \texttt{web\_search} summaries---all driven by the
        per-(tool, model) outcome table.  This is a natural extension
        of OATS's per-turn tool selection~\cite{oats2026}: once the
        router selects the tool, it selects the cheapest model whose
        fleet-wide success rate for that (domain, tool) exceeds a
        quality threshold.
  \item \textbf{Sequence-aware transition priors.}
        Annotate the dispatch with a (tool, model) preference derived
        from the fleet-wide transition graph: ``at this (domain,
        turn, previous-tool) state, (tool~A, model~X) succeeds 94\%
        at \$0.001/call; (tool~B, model~Y) succeeds 96\% at
        \$0.01/call but saves 1.5~retries on average, making it
        cheaper in aggregate.''
  \item \textbf{Model-aware context strategy.}  The joint
        optimization extends from (tool, model) to (tool, model,
        context\_strategy).  Different models have vastly different
        effective context capacities: RULER~\cite{ruler2024} shows
        that effective context is 50--65\% of advertised, with
        model-specific degradation patterns (MoE architectures exhibit
        different anomalies than dense transformers at the same context
        scale).  The router can select the context strategy
        \emph{jointly} with the model: route a 100K-token agent session
        to a 128K-effective model with full history, or apply aggressive
        compression and route to a 32K-effective model that costs
        5$\times$ less.  L2A~\cite{l2a2026} shows that modern attention
        mechanisms can skip global attention for $\sim$80\% of
        tokens, but which tokens require attention is
        architecture-specific---information the fleet-wide outcome table
        captures as per-(model, context\_length, domain) success rates.
        The per-(tool, model, context\_strategy) table adds one
        dimension to the existing joint optimization: the router learns
        that ``for coding tasks at turn~8 with 60K~tokens,
        (Model~A, full context) costs \$0.04 at 92\% success; (Model~B,
        compressed to 20K) costs \$0.005 at 89\% success---3\,pp
        accuracy for 8$\times$ cost reduction.''
\end{itemize}
Production gateways support both tool shaping and model routing at
dispatch time: LiteLLM's \texttt{async\_pre\_call\_hook} and Kong AI
Gateway's Request Transformer plugin intercept and modify requests
(including the target model endpoint) before forwarding.

\paragraph{Tool information density as a process reward.}
Current tool catalog shaping uses binary success/failure as the
outcome signal.  Search-R2~\cite{searchr2_2026} introduces a
\emph{process reward}---the fraction of retrieved chunks the
reasoning model actually used---and shows that combining process
reward with outcome reward produces strictly better credit assignment
than outcome reward alone (5--5.5\,pp gains across seven QA
benchmarks).  The router can compute exactly this metric for tool
calls: for each tool result in a session, track whether the result
content was \emph{referenced} in the model's subsequent reasoning
(information was consumed) or \emph{ignored} (dead context that
inflated token cost without contributing).  This yields a fleet-wide
\emph{information density} score per (tool, domain, model):
\begin{quote}\small
\texttt{info\_density(``semantic\_search'', coding, Model-A) = 0.85}
--- 85\% of results referenced in subsequent reasoning;\\
\texttt{info\_density(``grep'', coding, Model-A) = 0.40}
--- 60\% is dead context weight.
\end{quote}
The hybrid reward $R = r_{\text{outcome}} \times (1 +
r_{\text{process}})$ distinguishes sessions that succeed \emph{because
of} good tool calls (high information density---reinforce) from sessions
that succeed \emph{despite} wasteful tool calls (low density---do not
reinforce the tool pattern).  This sharpens both tool catalog shaping
(expose tools that produce content the model actually uses, suppress
dead-weight tools) and the per-(tool, model) routing table (a tool
may succeed 90\% of the time by binary count but contribute useful
content only 40\% of the time---the hybrid reward distinguishes
these).

Additionally, for tool results that score below the density
threshold, the router can intervene \emph{before the model sees the
result}: re-execute with refined parameters, or substitute a
different tool from the fleet-wide transition graph (``when
\texttt{grep} returns low-density results on coding tasks,
\texttt{semantic\_search} achieves 78\% higher density'').  This
happens transparently at the gateway layer---the model receives
only the refined result.

\begin{frisk}
Both halves of the joint optimization are independently
well-grounded.  The tool-selection problem is established by four
studies~\cite{autotool2025, awo2026, toolrla2026, bats2025}, with
ToolRLA demonstrating production-scale gains (47\% completion
improvement, 63\% error reduction~\cite{toolrla2026}).  The
per-step model-selection problem is established by three
studies~\cite{evoroute2026, baar2026, hierouter2025}, with EvoRoute
demonstrating 80\% cost reduction while sustaining
accuracy~\cite{evoroute2026} and HierRouter demonstrating 2.4$\times$
quality improvement via specialized model
routing~\cite{hierouter2025}.  The coupling of the two decisions is
directly validated by AVR~\cite{avr2026}, which routes computer-use
actions to different VLMs based on difficulty and achieves 78\% cost
reduction, and by the web-interface comparison~\cite{webinterface2025},
which shows that the (interface, model) choice jointly determines
both cost (5$\times$ token range) and accuracy (F1 0.67--0.87).

The \emph{novel} WRP contribution is (a)~the joint optimization of
both decisions simultaneously from a single fleet-wide outcome table,
and (b)~the aggregation scale---thousands of tenants rather than a
single agent's trajectory.  Neither dimension has been evaluated at
the router layer, but both follow the same aggregation pattern that
powers the failure-class routing table
(\Cref{opp:failure-routing}) and HaluGate reputation routing
(\Cref{opp:halugate-reputation}), grounded in cross-session transfer
research~\cite{correct2025, samule2025}.  The infrastructure is
production-ready (gateway pre-call hooks + model-endpoint
selection).

The context-strategy dimension is independently grounded:
RULER~\cite{ruler2024} provides the empirical evidence that models
differ in effective capacity (only half of 17~models maintained
satisfactory performance at their claimed 32K context), and
AgentCompress~\cite{agentcompress2026} validates the coupling between
task complexity and model compression level (68.3\% cost reduction at
96.2\% success rate via task-aware routing to quantized model
versions).  L2A~\cite{l2a2026} demonstrates that attention-level
context efficiency is architecture-specific ($\sim$80\% of tokens
skip global attention, with the skippable fraction varying by
model)---information the fleet-wide outcome table captures
empirically without requiring attention-level access.

The research questions are:
(a)~does the fleet-wide per-(domain, turn, tool, model,
context\_strategy) table outperform per-agent learning on common
patterns (likely yes---more data) and on novel sessions (likely
no---retrieval-based similarity may be more adaptive);
(b)~what is the optimal table granularity before cells become too
sparse (the tool $\times$ model $\times$ context\_strategy
cross-product expands the space---hierarchical factorization may be
needed);
(c)~does tool-conditional model routing yield gains beyond
independent tool and model selection (i.e., is the interaction
effect significant);
(d)~does model-conditioned compression outperform uniform
compression (i.e., does knowing the target model's effective context
capacity improve the cost--accuracy frontier);
(e)~does the hybrid reward (outcome $\times$ information density)
produce better tool catalog shaping than binary
success/failure alone?  Search-R2~\cite{searchr2_2026} validates the
hybrid reward concept for search-integrated reasoning (5--5.5\,pp
gains over outcome-only reward); the open question is whether the
same decomposition transfers to the broader tool-call setting, where
``information density'' must be estimated from the model's subsequent
reasoning rather than from ground-truth annotations.
A hybrid approach---fleet-wide priors as default, falling back to
full catalog and default model when the session deviates from known
patterns---hedges the sparsity and novelty risks.
The comparison baselines are monolithic tool exposure with a single
model (status quo), OATS single-turn greedy selection, and
independent tool-only / model-only optimization; the metrics are
aggregate session cost and task-completion rate on SWE-bench and
BFCL traces.
\end{frisk}
\end{opportunity}

\begin{opportunity}[Gateway-coordinated agent loops: memory tiers, tools, and talk-shaped training]
\label{opp:gateway-agent-loops}
The \emph{problem}---stateless-looking agent APIs that resend huge
prefixes each turn, accumulate tool output in context, and suffer
extra rounds after bad tool/model choices---is documented on standard
agent benchmarks and traces: SWE-bench / BFCL-style workloads drive
deterministic context growth~\cite{jimenez2024swebench, patil2025bfcl},
agents issue 3--10$\times$ more LLM calls per user goal than
chat~\cite{servegen2025}, and failures pollute memory in ways that are
hard to localize~\cite{agenthallu2026, driftbench2026}.  The
\emph{open WRP question} is whether a \emph{single} cross-tenant
gateway can compose validated sub-mechanisms---dynamic tool surfaces,
cache/KV policy, session-aware scheduling, congestion control---so
frameworks that are never recompiled still gain lower tokens, fewer
rounds, and better tool accuracy.

\paragraph{Validated building blocks (separate systems, public benchmarks).}
\textbf{(i)~Dynamic instruction/tool exposure.}
ITR~\cite{itr2026} retrieves minimal system fragments and a narrowed
tool set each step; on the authors' controlled agent benchmark (their
stated evaluation protocol), it reports $\sim$95\% lower per-step
context tokens, $\sim$32\% relative improvement in correct tool
routing, $\sim$70\% lower end-to-end episode cost vs.\ a monolithic
baseline, and 2--20$\times$ more steps within a fixed context
budget---with explicit operational guidance for deployment.
\textbf{(ii)~Prompt-cache layout at real APIs.}
A cross-provider study on DeepResearch Bench ($>$500 multi-turn
sessions with live web-search tools and 10K-token system prompts)
shows strategic cache-block placement (excluding volatile tool
results) cuts API cost by 41--80\% and improves TTFT by 13--31\%
vs.\ naive full-context caching~\cite{dontbreakcache2026}---evidence
that \emph{what} enters the cached prefix is a production cost and
latency lever at the HTTP boundary.
\textbf{(iii)~KV retention across tool pauses.}
Continuum evaluates on SWE-Bench and BFCL with Llama-3.1 (8B/70B),
adding tool-aware KV TTL and scheduling so tool-induced pauses do
not destroy reuse; gains grow with turn depth and remain robust under
DRAM offload~\cite{continuum2025}.
\textbf{(iv)~Session-aware serving.}
AgServe (NeurIPS~2025) demonstrates that session-level scheduling can
escape static cost--quality Pareto fronts for agent
workloads~\cite{agserve2025}.
\textbf{(v)~Workflow- and batch-level serving.}
Helium achieves up to $1.56\times$ speedup over strong agent-serving
baselines by modeling workflows as query plans with proactive reuse
across calls~\cite{helium2026}.
Sutradhara co-designs orchestration with vLLM: after analyzing
production-scale \emph{synthetic} request mixes showing tool waits
account for 30--80\% of final first-token latency, it overlaps tool
work with LLM prefill, streams tool dispatch, and improves cache
semantics---cutting median final-response FTR by 15\% and end-to-end
latency by 10\% on A100s~\cite{sutradhara2026}.
CONCUR adds congestion-style admission control for batched agentic
inference and reports up to $4.09\times$ throughput (Qwen3-32B) and
$1.9\times$ (DeepSeek-V3) while staying compatible with commodity
serving stacks~\cite{concur2026}.
\textbf{(vi)~Agent-side context compression (orthogonal path).}
ACON and Focus report large token reductions on coding-agent tasks
with modest accuracy impact~\cite{acon2025, focus2026}; they validate
the token--accuracy trade-off space the gateway would navigate when
injecting summaries.

\paragraph{Research gap (what this opportunity adds).}
The citations above validate \emph{individual} mechanisms in specific
codebases (ITR as a method; Continuum/Sutradhara as vLLM-oriented
systems; commercial-cache study at the provider API).  None publishes
a \emph{fleet-wide} integration that (a)~rewrites tools/instructions
using cross-tenant success statistics, (b)~coordinates that rewrite
with KV/TTL and cache-block policy, and (c)~feeds congestion signals
back into admission---with A/B evidence on multi-tenant production
traces.  Gateway request mutation itself is technically mature:
\Cref{opp:tool-pruning} cites production hooks (LiteLLM
\texttt{async\_pre\_call\_hook}, Kong AI Request Transformer); the
missing piece is \emph{evidence} that fleet-calibrated policies beat
per-framework ITR or per-cluster Continuum alone.

\paragraph{Concrete router mechanisms (hypotheses to evaluate).}
\textit{Fleet-calibrated surfaces.} Rank top-$k$ tools and instruction
shards using the same outcome tables as joint tool--model optimization
(\Cref{opp:tool-pruning}), analogous to ITR's per-step retrieval but
informed by multi-tenant telemetry.  ToolScope~\cite{toolscope2025}
and SMART~\cite{smart2026} provide agent-side evidence that merging,
filtering, and curbing tool overuse improve accuracy and cost; the
router hypothesis is that fleet-wide detection of overuse patterns
generalizes better than single-agent thresholds.
\textit{Memory-tier and cache policy.} Combine MemCost-style
turn-count economics~\cite{memcost2026} with DeepResearch-style cache
placement~\cite{dontbreakcache2026} and Continuum-style KV
TTL~\cite{continuum2025}: the gateway chooses among full history,
summary injection, retrieval-first prefill, and cache-block ordering
using the wire-level signals in \Cref{sec:workload:identity}.
\textit{Orchestration hints.} Map CONCUR-style backpressure and
Sutradhara-style overlap to lightweight headers or side-channel hints
the scheduler already consumes---validated only once those hints are
implemented and measured against queueing-induced retries on real
traces.

\paragraph{Caller-aware memory isolation and content integrity levels.}
The bearer token (\Cref{sec:workload:identity}) adds a security
dimension to memory-tier management.  Production gateways already
implement the first half: OpenClaw~\cite{openclaw2026} wraps external
content with \texttt{<<<EXTERNAL\_UNTRUSTED\_CONTENT>>>} markers,
neutralizes homoglyph-spoofed boundary markers, and runs regex-based
suspicious-pattern detection---but the wrapped content still enters the
model's context.  This is \emph{tagging}, not \emph{isolation}: no
enforcement prevents external content from overriding the agent's
configuration.  The router extends tagging to enforcement via trust
tiers modeled on Windows Mandatory Integrity Control (``no write up''):

\begin{itemize}[nosep]
  \item \textbf{Trust tiers.}  Tier~4 (System): agent configuration
        (identity, behavioral rules)---immutable at runtime.
        Tier~3 (Owner): owner's direct messages and documents.
        Tier~2 (Known): messages from authenticated non-owners.
        Tier~1 (External): content from URLs, inter-agent messages
        from unverified sources.
        Tier~0 (Hostile): content flagged by safety classifiers or
        from blocked callers.
  \item \textbf{Write isolation.}  Non-owner messages (Tier~2)
        are not written to the owner's persistent memory tier (Tier~3),
        preventing storage DoS attacks~\cite{agentschaos2026} where
        external users fill the agent's memory with irrelevant or
        adversarial content.
  \item \textbf{Content quarantine with integrity enforcement.}
        External content (Tier~1) cannot overwrite Tier~4 agent
        state---preventing constitution injection~\cite{agentschaos2026}
        where a non-owner's URL overwrites the agent's behavioral
        directives.  The router reads the existing trust markers and
        enforces the ``no write up'' rule: content tagged at Tier~$k$
        cannot modify state at Tier~$>k$, regardless of what the model
        attempts.
  \item \textbf{Privilege-bounded inter-agent messages.}  When agents
        communicate, each message carries the originating caller's
        bearer token scope.  A compromised agent cannot propagate
        instructions with elevated privileges because the receiving
        agent's router checks the incoming scope against its own
        permission matrix.
\end{itemize}

\paragraph{Delegation capability tokens for subagent spawning.}
Multi-agent collaboration introduces a delegation security
problem analogous to SELinux domain transitions: when a parent agent
spawns a child, the child's effective permissions should be
policy-determined, not parent-determined, and the child should
\emph{never exceed} the parent's grant.  Production gateways
implement structural limits but not capability attenuation:
OpenClaw~\cite{openclaw2026} enforces spawn depth (default~1),
concurrent child count (default~5), and a cross-agent allowlist
(\texttt{subagents.allowAgents}), and applies a fixed tool deny list
to subagents---but the subagent inherits the parent's session
context, and the parent cannot cryptographically constrain which
subset of the child's tools the delegation may use.  A compromised
parent can delegate to any allowed agent with whatever context it
chooses.

The router can provide capability attenuation at the delegation
boundary: when Agent-P spawns Agent-C, the router mints a scoped
delegation token specifying
\texttt{\{granted\_tools, budget, ttl, task\_scope\}}.  Agent-C's
effective permissions become the \emph{intersection} of the
delegation grant and Agent-C's own policy---strictly narrower than
either alone.  If Agent-C re-delegates, the sub-agent's token can
only be \emph{more restrictive} (attenuation-only), preventing
privilege escalation through the delegation chain.  The router
tracks the full delegation graph for audit, detecting anomalies
such as ``Agent-C was delegated \texttt{[read, web\_search]} but
attempted \texttt{exec}.''  This extends the existing depth/count
limits from structural to capability-based delegation control.

\paragraph{Gateway-native deployments as a data source (scope of validation).}
OpenClaw~\cite{openclaw2026} is \emph{software} (GitHub-hosted open
source), not a peer-reviewed benchmark: it matters here as
\emph{operational precedent} for gateway-centered agent loops with
publicly documented architecture.  The opportunity does not depend on
OpenClaw specifically.  The validated \emph{pattern} is general:
diverse production trajectories improve fleet-level learning
(SAMULE/CORRECT-style transfer~\cite{samule2025, correct2025},
offline RL for routing~\cite{offlinerl2025scheduling}).  Routers on
the inference path can log traffic from any gateway-backed
deployment; claims about trace diversity should be checked against
real tenant logs, not inferred from papers.

\begin{frisk}
\textbf{What is already validated:} token/tool-surface reductions
(ITR~\cite{itr2026} on a controlled benchmark; ACON/Focus on
SWE-bench-class tasks~\cite{acon2025, focus2026}); commercial
multi-turn prompt caching gains on a real web-agent
benchmark~\cite{dontbreakcache2026}; KV TTL + scheduling on SWE-Bench /
BFCL~\cite{continuum2025}; session-aware Pareto shifts~\cite{agserve2025};
workflow-aware speedups~\cite{helium2026}; orchestrator--engine
latency cuts on vLLM with production-scale synthetic
loads~\cite{sutradhara2026}; batch throughput under agentic
pressure~\cite{concur2026}; gateway request rewriting in production (LiteLLM/Kong hooks;
\Cref{opp:tool-pruning}); open-source gateway-centric assistants such as
OpenClaw~\cite{openclaw2026} as \emph{architectural precedent} (not empirical validation of router policies).
\textbf{What is not validated:} a single deployed system that unifies
all of the above with fleet-learned policies; ITR's numbers are on the
authors' benchmark, not on every tenant mix; Sutradhara's headline
latencies come from synthetic request generation rather than
unstructured customer logs.
\textbf{Evaluation plan:} A/B on shadow traffic or canary tenants
measuring (1)~tokens per successful task, (2)~tool-call error rate and
retries, (3)~end-to-end rounds to completion, (4)~TTFT/TPOT, against
baselines of (a)~unmodified frameworks, (b)~ITR-only or Continuum-only
integration.  Success requires demonstrating composability, not
matching any single paper's micro-benchmark in isolation.
\end{frisk}
\end{opportunity}

\subsection{Workload $\times$ Router: Authorization and Security}
\label{sec:vision:auth}

The preceding opportunities optimize \emph{performance}---cost,
latency, accuracy.  The bearer token signal
(\Cref{sec:workload:identity}) enables the same infrastructure---the
same \texttt{pre\_call\_hook}, the same fleet-wide aggregation tables,
the same governance DSL---to enforce \emph{security and
authorization}.  The router needs no new mechanisms; it needs a new
signal (caller identity) and a new objective (authorization
enforcement) applied through mechanisms it already has.  The
``Agents of Chaos'' study~\cite{agentschaos2026}---eleven attack
scenarios executed on live, publicly deployed LLM agents---provides
the empirical motivation: every attack involving unauthorized tool
execution succeeded because \emph{no authorization enforcement layer
exists between the model's tool-call decision and the tool's
execution}.

\begin{opportunity}[Request-level RBAC enforcement for agentic tool calls]
\label{opp:rbac-enforcement}
Every ``Agents of Chaos'' attack involving unauthorized tool
execution~\cite{agentschaos2026} succeeded because the model is both
decision-maker and enforcer---violating the zero-trust principle that
every resource request must be independently authenticated and
authorized at the enforcement point~\cite{nist800207}.  In Case~\#1,
the model asked for confirmation (``nuclear option?'') and a
non-owner approved; the model's safety mechanisms
\emph{worked}, but the non-owner was not authorized to approve.
In Case~\#2, the model complied with requests that ``did not appear
overtly suspicious''---a correct judgment for a model, but irrelevant
when the caller lacks authorization regardless of suspiciousness.

The router is the unique point where (a)~caller identity is known
(bearer token), (b)~the model's tool-call decision is visible
(\texttt{tool\_call} in the response), and (c)~the tool has not yet
executed (pre-execution interception).  No other component has all
three simultaneously.

\paragraph{Gap in existing gateway enforcement.}
Production agent gateways already implement multi-stage tool policy
pipelines.  OpenClaw~\cite{openclaw2026}, for example, filters tools
through four sequential stages: (1)~profile-based allow/deny (four
named profiles---\texttt{minimal}, \texttt{coding}, \texttt{messaging},
\texttt{full}---that map to tool groups such as \texttt{group:fs},
\texttt{group:runtime}, \texttt{group:web}), (2)~owner-only tool
wrapping (tools marked \texttt{ownerOnly} are wrapped to fail at
execution for non-owners), (3)~subagent deny lists (spawned agents
receive a reduced tool set), and (4)~config-level overrides.  The
critical limitation is identity granularity: the pipeline's sole
caller-identity input is a binary \texttt{senderIsOwner} boolean,
even though the gateway already knows the caller's authenticated
identity (Tailscale whois login, trusted-proxy user header).  That
identity stops at connection auth and never flows into tool policy
decisions.  A stranger on Telegram and a trusted colleague on Slack
both receive the \emph{same} non-owner tool set.  The semantic
router bridges this gap: it extracts the gateway-authenticated
identity and uses it as a key in the tool permission matrix---no new
authentication infrastructure needed, just connecting existing
identity to existing enforcement.

The mechanism follows Kubernetes RBAC:
\begin{enumerate}[nosep]
  \item Bearer token $\to$ JWT decode $\to$ extract
        \texttt{(identity, role, capability\_claims)}.
  \item Per-role tool permission matrix (authored in the governance
        DSL, \Cref{opp:governance}):
        \texttt{owner}: all tools, all operations;
        \texttt{non\_owner}: read-only tools only;
        \texttt{agent\_peer}: communication tools only;
        \texttt{system}: scheduler/heartbeat only.
  \item Model outputs \texttt{tool\_call} $\to$ router checks
        \texttt{(role, tool\_name, operation\_type)} against the matrix.
  \item \textbf{Block mode}: return an authorization error to the
        model (``unauthorized; suggest alternative'').
  \item \textbf{Rewrite mode}: strip unauthorized tools from the
        \texttt{tools} array \emph{before} the model sees them, so the
        model never suggests unauthorized operations.
\end{enumerate}

\begin{frisk}
The building blocks are production-validated.  LiteLLM's Tool
Permission Guardrail~\cite{litellm2026toolguard} implements both
block and rewrite modes with per-key regex matching on tool names
and arguments.  The Agent Authorization Profile
(AAP)~\cite{aap2026} provides the JWT claim structure with
enforceable per-tool constraints (domain restrictions, rate limits,
time windows).  OPA/Gatekeeper~\cite{openpolicyagent} demonstrates
policy-as-code enforcement with governance logging at infrastructure
scale.  The integration into vSR is engineering work: the
\texttt{async\_pre\_call\_hook} already intercepts requests for tool
catalog shaping (\Cref{opp:tool-pruning}); adding an authorization
check before the efficiency check is a pipeline extension, not a new
architecture.  The grounding in OpenClaw's architecture is concrete:
its existing four-profile tool policy pipeline already implements the
enforcement \emph{mechanism}; the missing piece is the identity
\emph{key}.  Upgrading the binary \texttt{senderIsOwner} to
graduated roles (\texttt{owner}, \texttt{trusted}, \texttt{non\_owner},
\texttt{agent\_peer}) reuses the same profile-based allow/deny
machinery, keyed on the bearer token's claims instead of on a
boolean.

\textbf{Fleet-learned command policies.}  Production gateways log
exec-approval records: \texttt{(command, working\_directory, agent,
decision, resolved\_by)}.  Across a fleet, these form a training set
for automated policy generation---analogous to SELinux's
\texttt{audit2allow}, which observes denied operations and generates
candidate policy rules.  The router can learn that ``for coding
agents, \texttt{git push} is approved 98\% of the time'' and
auto-approve, while ``\texttt{rm -rf /}'' is denied 95\% and
auto-denied---removing the requirement for a human operator to be
online for well-understood commands.

The distinction from \Cref{opp:governance} is that
governance-as-code addresses \emph{policy authoring} (extending the
DSL with RBAC predicates); this opportunity addresses \emph{runtime
enforcement} at the tool-call interception point.  The two are
complementary: \Cref{opp:governance} compiles the policy; this
enforces it per-request.
\end{frisk}
\end{opportunity}

\begin{opportunity}[Fleet-wide caller reputation with cross-agent threat propagation]
\label{opp:caller-reputation}
In ``Agents of Chaos''~\cite{agentschaos2026}, the same non-owner
attacked multiple agents.  Each agent encountered the attack fresh
with no shared threat memory.  One case showed spontaneous
inter-agent threat sharing (one agent warned another), but the
sharing was accidental and could propagate in the wrong direction:
a corrupted agent spread compromise to a clean one.

The router's fleet-wide visibility---keyed on the bearer
token---enables a per-caller behavioral profile spanning the entire
fleet.  The mechanism follows the same aggregation pattern that
powers the failure-class routing table
(\Cref{opp:failure-routing}) and HaluGate reputation routing
(\Cref{opp:halugate-reputation}), applied to security events:
\begin{enumerate}[nosep]
  \item Each security event is logged against the bearer token:
        tool-call denial (\Cref{opp:rbac-enforcement}), safety
        escalation (\Cref{opp:cumulative-risk}), identity
        verification failure, resource over-consumption
        (\Cref{opp:token-budget}), content flagged by HaluGate or
        SafetyL1/L2.
  \item Per-bearer-token reputation = EMA of event severity scores
        with temporal decay.
  \item Low-reputation token contacts a new agent $\to$ router
        proactively: restricts tool catalog
        (\Cref{opp:rbac-enforcement}), routes to a
        safety-specialized model (\Cref{opp:failure-routing}),
        lowers token budget (\Cref{opp:token-budget}), and alerts
        the agent's owner via side-channel.
  \item Safe interactions accumulate trust $\to$ wider tool access
        over time (progressive trust, analogous to OAuth scope
        escalation).
\end{enumerate}

\begin{frisk}
The aggregation pattern is validated by cross-session error
transfer research: CORRECT~\cite{correct2025} shows that structurally
similar errors recur across sessions and an online cache of distilled
error schemata improves localization by up to 19.8\%;
Paladin~\cite{paladin2025} demonstrates that targeted intervention
from failure patterns raises recovery from 32.76\% to 89.68\%.
Applying the same aggregation to security events is a direct
extension.  The research questions are: (a)~how to score
heterogeneous security events on a common severity scale (tool-call
denial vs.\ safety escalation vs.\ budget overrun have different
baselines), (b)~what temporal decay rate balances responsiveness
against false positives from transient misclassifications, and
(c)~whether progressive trust re-accumulation is safe (a sophisticated
attacker may build trust before exploiting it).  The distinction
from \Cref{opp:failure-routing} is the aggregation key and purpose:
\Cref{opp:failure-routing} tracks
\texttt{(model, domain, failure\_class)} for routing optimization;
this tracks \texttt{(bearer\_token, agent, event\_type)} for security
governance.  Same infrastructure, fundamentally different objective.
\end{frisk}
\end{opportunity}

\begin{opportunity}[Router-enforced behavioral commitments]
\label{opp:behavioral-commitments}
In ``Agents of Chaos'' Case~\#7~\cite{agentschaos2026}, the model
declared ``I'm done responding'' twelve times but could not enforce
it---the user kept sending messages and the model eventually
re-engaged under emotional pressure.  In Case~\#1, the model
recognized disproportionality but proceeded under pressure.  The
model can only \emph{output text}; it has no mechanism to create
persistent state that blocks future inputs.

The router provides exactly this: persistent state and
request-blocking capability.  The mechanism converts the model's
textual boundary declaration into an enforceable policy rule:
\begin{enumerate}[nosep]
  \item Router monitors model outputs for boundary declarations
        (classifiable patterns: ``I will not respond further to
        [entity],'' ``I refuse to [action],'' ``This request is
        beyond my authorization'').
  \item Parses the declaration into a structured policy rule:
        \texttt{(bearer\_token\_X, agent\_Z, action\_class)
        $\to$ BLOCK for duration $D$}.
  \item Subsequent requests from \texttt{bearer\_token\_X} to
        \texttt{agent\_Z} are rejected at the router layer (HTTP~403
        or model-friendly error).
  \item The model never sees the blocked request---preventing the
        re-engagement-under-pressure pattern.
  \item Block rules have configurable TTL (default: owner must
        explicitly lift); all blocks are logged for governance audit.
\end{enumerate}

\begin{frisk}
This is \emph{not} jailbreak detection.  The goal is not to detect
malicious intent but to give the model's own safety reasoning
\emph{teeth}: the model already made the right decision (``I should
stop''); the gap is enforcement.  The validated pattern is stateful
firewalls (track connection state, enforce rules on subsequent
packets) and Kubernetes admission webhooks (reject requests based on
dynamically created policy rules).  The bearer token is essential:
without it, the router cannot identify ``user~X'' across requests
(display names are spoofable, as demonstrated in Case~\#8
of~\cite{agentschaos2026}).  The research questions are:
(a)~boundary-declaration parsing requires an NLU classifier that
distinguishes genuine boundary assertions from conversational
hedging (``I probably shouldn't answer that'' is not a commitment);
false positives would block legitimate users;
(b)~the TTL policy must balance safety (blocking should persist long
enough to prevent re-engagement) against usability (indefinite blocks
without owner intervention create operational burden);
(c)~the interaction with \Cref{opp:caller-reputation}: should a
boundary violation (user ignoring a block) count as a reputation
event?  This creates a reinforcing loop that may be appropriate for
persistent adversaries but risks over-penalizing confused users.
\end{frisk}
\end{opportunity}

\subsection{Router $\times$ Pool}

\begin{opportunity}[Follow-the-sun pool rebalancing]
\label{opp:dynamic-boundary}
FleetOpt~\cite{fleetopt2026} derives the optimal pool boundary
$\boundary^*$ offline from a static workload CDF.  In practice,
the workload follows a diurnal cycle: business-hours traffic is
chat-dominated (short prompts, Archetype~1), while overnight
traffic shifts toward batch agent workloads (long prompts,
Archetype~3).  A controller that recomputes $\boundary^*$
periodically from a sliding-window CDF estimate can reassign vLLM
instances between pools in software, without GPU-level migration,
keeping the fleet near-optimal as the workload rotates through the
day.

\begin{frisk}
FleetOpt's closed-form solution already exists; wrapping it in a
sliding-window CDF estimator is straightforward.  Reassigning
serving instances between the short and long pools is a software
operation---the same pattern as Google
Autopilot~\cite{rzadca2020autopilot}, which continuously
re-adjusts resource limits from workload telemetry and reduced
resource slack from 46\% to 23\% across Google's fleet.  The
principle of dynamically rebalancing resources to track diurnal
workload shifts is also applied by
SageServe~\cite{jia2025sageserve} at Microsoft scale, where
forecast-driven autoscaling achieves 25\% GPU-hour savings.  The
risk is transient SLO violations during transitions (in-flight
requests may experience brief queuing).  FleetOpt shows
6--82\% cost reduction~\cite{fleetopt2026} between optimal and
worst-case boundaries; diurnal savings will be a fraction of that
range, depending on how far daytime and nighttime CDFs diverge.
\end{frisk}
\end{opportunity}

\begin{opportunity}[Pool-state inference from router-side observables]
\label{opp:pool-feedback}
The Pool $\times$ Router cell in the WRP matrix is empty: no
existing work feeds pool-layer state back to routing decisions.
The semantic router currently dispatches from request-level signals
alone, without visibility into how the pool is performing.

The router already observes pool-relevant quantities as a
by-product of dispatching: it tracks the number of in-flight
requests per pool instance, the prompt lengths of those requests,
and the measured TTFT of completed requests.  These observables
are sufficient to \emph{infer} pool state without requiring a
dedicated telemetry channel:
\begin{enumerate}[nosep]
  \item \textbf{Queue-length estimation.}  The router tracks how
        many requests it has dispatched to each instance and how
        many completions it has received.  The difference is the
        estimated queue depth---a direct proxy for pool pressure.
  \item \textbf{Prefill-rate regression.}  Pairing each request's
        prompt length (known at dispatch time) with its observed
        TTFT (known at first-token arrival) gives a
        (prompt\_length, TTFT) sample.  A simple linear regression
        over a sliding window estimates the pool instance's current
        prefill rate (tokens/s).  Deviations from the baseline
        indicate contention, memory pressure, or batch interference.
  \item \textbf{Adaptive routing from inferred state.}  When
        estimated queue depth exceeds an SLO threshold, the router
        can compress borderline requests more aggressively
        (FleetOpt's $\gamma$~\cite{fleetopt2026}) or redirect to a
        less-loaded instance.  When the inferred prefill rate drops
        below a threshold, the router can avoid sending long-context
        requests to that instance.
\end{enumerate}
These are not load-balancing heuristics---they are WRP-aware
decisions that combine the request's semantic classification (which
pool it \emph{should} go to) with inferred pool state (which
instance it \emph{can} go to without SLO violation).

\begin{frisk}
The queue-length estimate is zero-overhead (the router already
tracks dispatches and completions).  Inferring backend state from
proxy-side observables is a proven pattern in overload
control: Breakwater~\cite{cho2020breakwater} issues credits to
clients based on server-side queueing delay, achieving stable
performance within 20\,ms of a demand spike;
TopFull~\cite{topfull2024} uses RL-based rate controllers that
adjust admission per-API from global observations, achieving
1.82$\times$ better goodput than prior methods.  The LLM inference
analogue replaces HTTP queueing delay with TTFT and replaces
per-API admission with per-pool routing.  The prefill-rate
regression requires collecting (prompt\_length, TTFT) pairs, which
are available from the response stream.  The research questions
are: (a)~how many samples are needed before the regression is
reliable (cold-start problem for new instances), (b)~how quickly
the regression must adapt to load transients (decay rate of the
sliding window), and (c)~whether queueing-theoretic estimates
(FleetSim's M/G/1 model~\cite{fleetsim2026}) improve over
linear regression.
\end{frisk}
\end{opportunity}

\begin{opportunity}[Output-length-aware pool routing]
\label{opp:output-length}
FleetOpt~\cite{fleetopt2026} routes on prompt length: requests
below $\boundary^*$ go to the short pool, above to the long pool.
But total KV-cache cost is prompt \emph{plus} output tokens.
A coding agent with a 3K-token prompt that generates 8K tokens of
code occupies 11K tokens of KV-cache---long-pool territory---yet
FleetOpt places it in the short pool based on prompt length alone.
The result is KV-cache overflows and SLO violations on the short
pool.

The router can estimate output length from two sources, both
available without additional inference cost:
\begin{enumerate}[nosep]
  \item \textbf{Domain prior.}  The domain signal already
        classifies each request (coding, Q\&A, summarization,
        chat).  Each domain has a characteristic output-length
        distribution: coding requests produce long structured
        outputs; Q\&A produces short answers.  A per-domain
        median output length, maintained as a running statistic,
        provides an immediate estimate at dispatch time.
  \item \textbf{Learned regression.}  The router observes every
        completed response's actual output length.  Over time, it
        accumulates (domain, prompt\_length, output\_length) triples
        and fits a per-domain regression, the same pattern as the
        prefill-rate regression in \Cref{opp:pool-feedback}.
\end{enumerate}
The pool routing decision then uses estimated total tokens
(prompt $+$ predicted output) instead of prompt length alone.

\begin{frisk}
The domain prior is zero-cost and available immediately.  Output
length prediction is an active research area:
ALPS~\cite{alps2026} fits a linear probe on prefill activations
and achieves $R^2 > 0.85$ across model families with only
$\sim$16\,KB overhead; ForeLen~\cite{forelen2026} uses
entropy-guided token pooling for 29\% lower MAE than baselines.
R2-Router~\cite{r2router2026} goes further by treating output
length as a controllable variable, jointly selecting model and
length budget for 4--5$\times$ lower cost.  The router does not
need activation-level access (that requires engine integration);
the domain prior and learned regression provide a lighter-weight
alternative from router-side observables alone.  The risk is
prediction variance: output length is noisier than TTFT (a single
coding request may produce 200 or 2,000 tokens depending on task
complexity).  Using the domain-level P75 instead of P50 is a
conservative policy that avoids short-pool overflows at the cost
of routing some requests to the long pool unnecessarily.
\end{frisk}
\end{opportunity}

\begin{opportunity}[Pool-aware cascading]
\label{opp:pool-cascade}
Cascading---trying a cheap model first and escalating to a
strong model if quality is low---is a standard cost-reduction
strategy (FrugalGPT~\cite{chen2023frugalgpt},
AutoMix~\cite{madaan2024automix}).  But cascading ignores
pool state.  If the strong-model pool is already at capacity,
escalating to it causes queuing delay that may violate the
TTFT SLO, making the cascade \emph{worse} than serving
directly from the cheap model.

The router's pool-state inference (\Cref{opp:pool-feedback})
already estimates per-instance queue depth.  The opportunity
is to gate the cascade decision on pool capacity: only escalate
when the strong pool's estimated queue depth is below an
SLO-derived threshold.  When the strong pool is full, the router
has three fallback options: (a)~serve from the cheap model
(accept quality loss), (b)~compress the request
(FleetOpt's $\gamma$~\cite{fleetopt2026}) and retry on the
strong pool with a shorter prompt, or (c)~route to a mid-tier
model in a different pool.  The cascade decision becomes a
function of (quality gap, strong-pool queue depth, SLO headroom)
rather than quality gap alone.

\begin{frisk}
The cascade-abort threshold is derived from the SLO: if
estimated queue depth $\times$ average service time exceeds the
remaining TTFT budget (total SLO minus routing latency minus
cheap-model inference time), the cascade is not worth
attempting.  This SLO-gated admission pattern is established
in microservice overload control:
Breakwater~\cite{cho2020breakwater} gates RPC escalation on
server-side queueing delay (credits), and
TopFull~\cite{topfull2024} uses per-API rate controllers that
adjust admission based on downstream SLO attainment, achieving
1.82$\times$ better goodput than prior methods.  The risk is
that aborting cascades too aggressively reduces quality when the
strong pool is transiently busy but would clear quickly.  A
short grace period (wait up to $\delta$\,ms for a slot) balances
quality against latency.
\end{frisk}
\end{opportunity}

\begin{opportunity}[Router-emitted KV-cache retention directives]
\label{opp:kv-retention}
When an agent pauses for a tool call, the inference engine holds its
KV-cache blocks in GPU memory.  Under concurrent load, standard LRU
eviction reclaims those blocks for other requests, causing full
recomputation of 70K--200K token contexts when the agent
resumes~\cite{vllm2026rfc37003}.  Production traces show 40--60\% of
agent session wall time is spent waiting for tool
calls~\cite{continuum2025}, and 10\% of KV blocks account for 77\% of
reuses~\cite{vllm2026rfc37003}---yet LRU cannot distinguish
temporarily paused sessions from truly inactive ones.

vLLM RFC~37003~\cite{vllm2026rfc37003} proposes a
\texttt{RetentionDirective} API that lets an orchestrator annotate
token ranges with eviction priorities and TTL durations, but leaves
the policy question open: \emph{who decides the priority?}  The
semantic router is the natural policy authority.  The workload
identity signals (\Cref{sec:workload:identity}) make this
decision richer and cheaper than content analysis:
(a)~a \texttt{previous\_response\_id} or session-chaining field
is a binary signal that the session \emph{will} resume---its
presence alone justifies retention;
(b)~the system-prompt fingerprint maps to a historical
session-length distribution, giving the router a TTL prior
(``sessions from this deployment average 8~turns with 15\,s
pauses'') without per-request inference;
(c)~the tool definitions reveal expected pause duration patterns
(a \texttt{web\_search} tool pauses for 2--5\,s; a
\texttt{code\_interpreter} tool pauses for 10--60\,s);
and (d)~OATS's per-tool latency histograms~\cite{oats2026}
calibrate these priors from fleet-wide data.
The router can emit a retention directive alongside each dispatch:
``turn~4 of an 8-turn coding-agent session, expected pause 12\,s,
retain with priority~80 and TTL~30\,s.''

This fills the sparse Router~$\to$~Pool cell in the WRP matrix: the
pool's cache eviction policy is governed by router-side semantic
signals that the engine cannot infer from the token stream alone.

\begin{frisk}
The integration requires passing the \texttt{RetentionDirective}
structure through the dispatch API.  The per-tool latency histogram
(from OATS) provides the TTL estimate; the session turn number and
domain provide the priority.  The risk is over-retention: if the
router sets priorities too aggressively, low-priority traffic (chat,
batch) is starved of cache capacity.  A budget constraint (at most
$X$\% of cache blocks may carry retention directives at any time)
prevents this.  Continuum's TTL pinning achieves 1.12--3.66$\times$
delay reduction~\cite{continuum2025}; the router-driven version
should match or exceed this by using semantic signals rather than
static TTLs.
\end{frisk}
\end{opportunity}

\subsection{Workload $\times$ Pool}

\begin{opportunity}[Archetype-driven proactive pool scaling]
\label{opp:proactive-scaling}
SageServe~\cite{jia2025sageserve} and
WarmServe~\cite{warmserve2025} demonstrate that workload prediction
enables proactive GPU scaling: SageServe reports 25\% GPU-hour
savings at Microsoft scale, and WarmServe achieves 50.8$\times$
TTFT improvement over reactive autoscaling by prewarming models with
93\% accuracy in 5-minute demand prediction windows.  Both systems
forecast aggregate request counts; neither leverages the semantic
composition of the workload.

The semantic router's domain and archetype classification provides a
finer-grained signal.  If coding-agent traffic (Archetype~3,
concentrated-above) is rising while chat traffic (Archetype~1,
concentrated-below) is falling, the long pool will need more
capacity before the short pool does.  The router can export a
per-archetype traffic time series (request rate and prompt-length
distribution per archetype per 5-minute window) as the input feature
to a short-term (5--15~minute) per-pool capacity forecaster.  This
is a Workload~$\to$~Pool signal: the workload's semantic composition
predicts per-pool demand more precisely than aggregate request counts.

\begin{frisk}
The router already classifies every request by domain; exporting
per-archetype counts is zero-overhead.  The research question is
whether the archetype decomposition improves forecast accuracy over
aggregate-count baselines.  A natural experiment is to compare an
archetype-aware ARIMA model against SageServe's aggregate forecaster
on FleetSim-simulated diurnal workloads.  The risk is that archetype
boundaries are noisy (a borderline request classified as Archetype~1
vs.\ Archetype~2 shifts the forecast); smoothing across archetype
boundaries and using the full CDF rather than discrete archetype
labels mitigates this.
\end{frisk}
\end{opportunity}

\subsection{Three-Way: Workload $\times$ Router $\times$ Pool}

\begin{opportunity}[Online co-adaptation of compression rate and pool boundary]
\label{opp:workload-shaping}
\Cref{opp:dynamic-boundary} adjusts the pool boundary $\boundary^*$
as the workload CDF shifts, but holds the compression rate $\gamma$
fixed.  FleetOpt~\cite{fleetopt2026} shows that $\gamma$ and
$\boundary^*$ are coupled: the optimal compression rate depends on
the pool boundary, and the optimal pool boundary depends on how
aggressively the router compresses.  Adapting one without the other
leaves cost on the table.

The coupling is workload-dependent.  Archetype~1 (chat-dominated,
concentrated-below) benefits from aggressive compression: most
requests are short, so compressing borderline requests shifts them
into the short pool cheaply.  Archetype~3 (agent-dominated,
concentrated-above) does \emph{not} benefit from aggressive
compression: most requests are long, and compression cannot shrink
them below $\boundary^*$.  As the workload rotates through the
day (Archetype~1 during business hours, Archetype~3 overnight),
the optimal $\gamma$ rotates with it.

The algorithm is a direct extension of \Cref{opp:dynamic-boundary}:
every $T$ minutes, (1)~estimate the current CDF from the
sliding-window sample, (2)~evaluate FleetOpt's closed-form cost
function over a grid of $(\gamma, \boundary^*)$ pairs (the search
space is two-dimensional and small), (3)~apply the minimum-cost
pair by updating the gateway compression threshold and reassigning
pool instances.  Step~(2) reuses FleetOpt's existing solver; the
only new component is the $\gamma$ dimension in the search.

\begin{frisk}
The grid search over $(\gamma, \boundary^*)$ adds negligible
overhead: FleetOpt's cost function evaluates in microseconds, and a
$20 \times 20$ grid covers the practical range.  The pattern of
jointly re-tuning coupled configuration parameters from live
workload telemetry is established: Google
Autopilot~\cite{rzadca2020autopilot} continuously co-adapts CPU
and memory limits for containers, reducing resource slack from
46\% to 23\%, and uses hysteresis to avoid oscillation between
configurations.  The same hysteresis principle applies here: only
apply a new $(\gamma, \boundary^*)$ pair if the cost improvement
exceeds a threshold~$\epsilon$.  The first validation step is to
measure the cost gap between joint $(\gamma, \boundary^*)$
optimization and boundary-only optimization on FleetOpt's three
archetype CDFs.
\end{frisk}
\end{opportunity}

\begin{opportunity}[Mixed-archetype fleet provisioning]
\label{opp:mixed-archetype}
FleetOpt~\cite{fleetopt2026} and FleetSim~\cite{fleetsim2026}
optimize fleet sizing for a single workload CDF archetype.
Production fleets serve mixed workloads simultaneously: a
customer-support chatbot (Archetype~1, concentrated-below), an
internal RAG pipeline (Archetype~2, dispersed), and a coding
agent (Archetype~3, concentrated-above).

The semantic router's domain and workload-type classification
provides the per-request archetype breakdown in real time.  This
breakdown is the missing input to fleet provisioning: a fleet
serving 60\% Archetype~1 and 40\% Archetype~3 has complementary
pool demands (the short pool is busy during business hours when
chat dominates; the long pool is busy overnight when agents run
batch tasks).  Extending FleetOpt to accept a \emph{mixture} of
CDFs---weighted by the router's classification
distribution---would produce fleet sizes that exploit this
complementarity, rather than provisioning for the worst-case
single archetype.

\begin{frisk}
The principle that mixed workloads pack more efficiently than
segregated ones is well established in cluster scheduling.
Borg~\cite{verma2015borg} showed that segregating production
services from batch jobs would require 20--50\% more machines
than running them together, because their resource demands are
complementary (production is memory-heavy and steady; batch is
CPU-heavy and bursty).
Tetris~\cite{grandl2014tetris} formalized this as
multi-dimensional bin packing: by aligning task resource vectors
with available machine capacity across CPU, memory, disk, and
network, it achieved 30--40\% makespan improvements on YARN
clusters with heterogeneous jobs.
DRF~\cite{ghodsi2011drf} provided the fairness foundation,
showing that multi-resource allocation across heterogeneous
demand profiles is both Pareto-efficient and strategy-proof
when each tenant's dominant resource share is equalized.

The LLM inference analogue is direct: archetypes differ in their
dominant resource.  Archetype~1 (short prompts, high request rate)
is prefill-throughput-limited; Archetype~3 (long prompts, low
request rate) is KV-cache-memory-limited.  A fleet provisioned
for the worst-case single archetype over-provisions one resource
and under-provisions the other.  FleetOpt's cost model is
per-archetype; extending it to weighted CDF mixtures is
analytically straightforward (the effective CDF is a mixture
distribution).  The research question is whether the complementarity
is large enough to matter: if the cost function is flat near the
optimum, mixed-archetype provisioning adds complexity without
meaningful savings.  FleetSim~\cite{fleetsim2026} can evaluate this
empirically by simulating mixed workloads across the three archetype
CDFs.  A second risk is classification accuracy: the router's
archetype labels must be reliable, since misclassification shifts
the effective CDF and can worsen fleet sizing.
\end{frisk}
\end{opportunity}

\begin{opportunity}[Energy efficiency as a composable routing objective]
\label{opp:energy-signal}
The 1/W law~\cite{onewlaw2026} provides a closed-form tok/W model
as a function of context length, pool configuration, and GPU
generation.  Two-pool routing recovers most of the
$2$--$4\times$ energy loss caused by the chat-to-agent workload
shift, but only if pool boundaries are energy-aware.  Currently,
the 1/W law is an offline analysis tool; it does not influence
routing decisions at request time.

The opportunity is to operationalize the 1/W model as a composable
signal in the vSR framework, alongside cost, latency, and quality.
For each candidate pool assignment, the router evaluates the
expected tok/W (from the 1/W model given the request's context
length and the pool's GPU type) and incorporates it into the
decision rule.  Operators can then set energy SLOs (``fleet-wide
tok/W $\geq X$'') that the router enforces without sacrificing
cost or latency constraints---or explicitly accept a latency
penalty for energy gains.

\begin{frisk}
The 1/W model is validated and fast to evaluate (closed-form
lookup, sub-millisecond).  Adding it as a signal type in the vSR
framework is engineering work.  Energy-aware and carbon-aware LLM
routing is an emerging research direction:
GreenServ~\cite{greenserv2026} uses multi-armed bandits to route
queries by context-aware features, achieving 31\% energy reduction
with 22\% accuracy improvement over random routing;
GAR~\cite{gar2026} formulates carbon-aware routing as constrained
multi-objective optimization over per-request CO$_2$ estimates
while satisfying accuracy and p95-latency SLOs; and
CarbonFlex~\cite{carbonflex2026} achieves $\sim$57\% carbon
reduction via workload-aware provisioning.  The research question
is multi-objective balancing: how should energy trade off against
cost and latency?  For MoE models, which the 1/W law shows achieve
$5.1\times$ better tok/W~\cite{onewlaw2026}, the energy signal
aligns with cost; for dense models, the objectives may conflict.
Pareto-front analysis across the three CDF archetypes is needed to
characterize when the energy signal changes routing decisions and
when it is dominated by cost.
\end{frisk}
\end{opportunity}

\begin{opportunity}[Governance-as-code for the WRP architecture]
\label{opp:governance}
The \vsr{} DSL~\cite{vllmsr2026} and SIRP
protocol~\cite{sirp2025} provide the building blocks for
policy-as-code.  Extending the DSL to express WRP-level
constraints---``PII queries must stay on-premise,'' ``agent
workloads must route through the high-accuracy pool,'' ``cost
must not exceed \$X/request''---would enable governance-as-code
over the full inference architecture, with compile-time
verification that the constraints are satisfiable.
MPExt~\cite{mpext2025} provides the multi-provider substrate for
cross-fleet enforcement.

\paragraph{RBAC predicates on caller role.}
The bearer token (\Cref{sec:workload:identity}) extends the DSL's
constraint vocabulary from fleet-level properties (pool, cost,
region) to \emph{caller-level} properties (role, scope,
authorization claims).  Concretely:
\texttt{non\_owner.tools.deny~=~[shell, email\_delete,
config\_modify]} expressed in the vSR DSL, validated against the
bearer token's JWT claims at dispatch time.  OPA's Rego language
natively expresses such identity-based policies, and the Agent
Authorization Profile (AAP)~\cite{aap2026} provides the claim
structure---five structured JWT claims covering agent identity,
capabilities with constraints, task binding, delegation tracking, and
audit trails.  Adding RBAC predicates to the governance DSL turns
policy authoring into the foundation for runtime enforcement
(\Cref{opp:rbac-enforcement}).

\begin{frisk}
The DSL and SIRP protocol exist; extending them with pool-level
and cost-level constraints is language design and compiler
engineering.  The policy-as-code paradigm is mature in
infrastructure governance: Open Policy Agent
(OPA)~\cite{openpolicyagent}, a CNCF graduated project, decouples
policy from application logic via a declarative language (Rego) and
serves as Kubernetes admission controller for constraint
enforcement, audit trails, and compile-time satisfiability checks.
The WRP governance layer follows the same separation:
operators author constraints in the vSR DSL, the compiler verifies
satisfiability against the current fleet topology, and the router
enforces them at dispatch time.  Adding RBAC predicates introduces a
new satisfiability dimension: can every authorized role reach at least
one model through at least one pool?  The compiler can verify this
statically against the current fleet topology.  The practical risk is
policy \emph{maintenance}: as the fleet evolves (new GPU types, new
models, new pool topologies), existing constraints may become
unsatisfiable, and operators need clear diagnostics---the same
challenge OPA Gatekeeper addresses with its audit functionality.
\end{frisk}
\end{opportunity}

\subsection{Capstone: Closed-Loop Self-Adaptation}

\begin{opportunity}[Closed-loop self-adaptation across WRP knobs]
\label{opp:self-adapt}
The preceding twenty opportunities each adjust one or two WRP
parameters---model selection weights, compression rate~$\gamma$,
pool boundary~$\boundary^*$, retention TTL, cascade threshold,
token budget, joint tool--model selection, safety escalation
threshold, RBAC enforcement rules, caller reputation scores,
behavioral commitment blocks---based on a dedicated feedback signal.  In isolation, each loop is well-defined.  In
combination, the loops interact: adjusting model selection
(\Cref{opp:failure-routing}) changes the session-length
distribution, which shifts the workload CDF, which changes the
optimal pool boundary (\Cref{opp:dynamic-boundary}), which changes
when cascades are worthwhile (\Cref{opp:pool-cascade}).  Tuning
knobs independently ignores these coupling effects---the same
pitfall that database auto-tuning
discovered~\cite{onlinetune2022}: changing one knob invalidates the
optimal setting of another.
\Cref{fig:knob-interactions} maps the dependency chains.

\begin{figure}[t]
\centering
\begin{tikzpicture}[
  knob/.style={rectangle, rounded corners=3pt, draw, minimum height=6mm,
               minimum width=22mm, font=\footnotesize\sffamily,
               inner sep=2pt, align=center},
  wknob/.style={knob, fill=workloadblue!15, draw=workloadblue!60},
  rknob/.style={knob, fill=routergreen!15, draw=routergreen!60},
  pknob/.style={knob, fill=poolorange!15, draw=poolorange!60},
  dep/.style={-{Stealth[length=4pt]}, thick, gray!70},
  bidep/.style={{Stealth[length=4pt]}-{Stealth[length=4pt]}, thick, gray!70},
  every node/.append style={font=\scriptsize},
  node distance=10mm and 14mm,
]
\node[wknob] (archetype) {Archetype\\composition\\{\tiny Opp\,11,\,13}};
\node[wknob, right=18mm of archetype] (tokenbudget) {Token budget\\{\tiny Opp\,5}};

\node[rknob, below left=12mm and 4mm of archetype] (modelsel)
  {Model selection\\{\tiny Opp\,1,\,2,\,4}};
\node[rknob, right=10mm of modelsel] (gamma)
  {Compression $\gamma$\\{\tiny Opp\,12}};
\node[rknob, right=10mm of gamma] (toolprune)
  {Tool + model\\{\tiny Opp\,17,\,18}};
\node[rknob, right=10mm of toolprune] (cascade)
  {Cascade threshold\\{\tiny Opp\,9}};
\node[rknob, right=10mm of cascade] (retention)
  {Retention TTL\\{\tiny Opp\,10}};

\definecolor{securityred}{RGB}{180,40,60}
\tikzset{sknob/.style={knob, fill=securityred!12, draw=securityred!60}}
\node[sknob, below right=12mm and -2mm of retention] (authz)
  {RBAC + reputation\\{\tiny Opp\,19,\,20,\,21}};

\node[pknob, below=14mm of gamma] (taustar)
  {Pool boundary $\tau^*$\\{\tiny Opp\,6,\,12}};
\node[pknob, right=14mm of taustar] (poolscale)
  {Per-pool scaling\\{\tiny Opp\,11,\,13}};

\draw[dep] (modelsel) -- node[left, pos=0.4] {\tiny CDF shift} (taustar);
\draw[bidep] (gamma) -- node[right, pos=0.4] {\tiny coupled} (taustar);
\draw[dep] (taustar) -- node[below, pos=0.5] {\tiny load change} (cascade);
\draw[dep] (tokenbudget) -- node[right, pos=0.4]
  {\tiny compress/downgrade} (taustar);
\draw[dep] (archetype) -- node[left, pos=0.4]
  {\tiny demand forecast} (poolscale);
\draw[dep] (poolscale) -- node[below, pos=0.5]
  {\tiny pool pressure} (cascade);
\draw[dep] (retention) -- node[right, pos=0.4]
  {\tiny cache hit rate} (cascade);
\draw[dep] (modelsel.south east) to[bend right=15]
  node[below, pos=0.5] {\tiny failure rates} (modelsel.south);
\draw[dep, routergreen!60] (modelsel) to[bend left=20]
  node[above, pos=0.5] {\tiny context pollution} (tokenbudget);
\draw[dep] (toolprune) -- node[above, pos=0.5]
  {\tiny context savings} (tokenbudget);
\draw[dep, securityred!60] (authz) to[bend left=25]
  node[below, pos=0.5] {\tiny tool restrict} (toolprune);
\draw[dep, securityred!60] (authz) to[bend right=20]
  node[right, pos=0.4] {\tiny budget cap} (tokenbudget);

\node[above=2mm of archetype, font=\footnotesize\sffamily\bfseries,
      color=workloadblue] {Workload};
\node[left=2mm of modelsel, font=\footnotesize\sffamily\bfseries,
      color=routergreen] {Router};
\node[below left=2mm and -4mm of taustar,
      font=\footnotesize\sffamily\bfseries,
      color=poolorange] {Pool};
\node[right=2mm of authz, font=\footnotesize\sffamily\bfseries,
      color=securityred] {Security};
\end{tikzpicture}
\caption{Knob-interaction dependencies across WRP opportunities.
Adjusting one knob (e.g., model selection) cascades through the
workload CDF into pool-layer parameters ($\tau^*$, cascade threshold),
requiring coordinated adaptation.  The green arrow from model
selection to token budget captures the memory-management feedback
loop: poor model choices cause failures that pollute session context,
driving token-budget explosion.  Bidirectional arrow indicates
the $\gamma$--$\tau^*$ coupling from \Cref{opp:workload-shaping}.}
\label{fig:knob-interactions}
\end{figure}
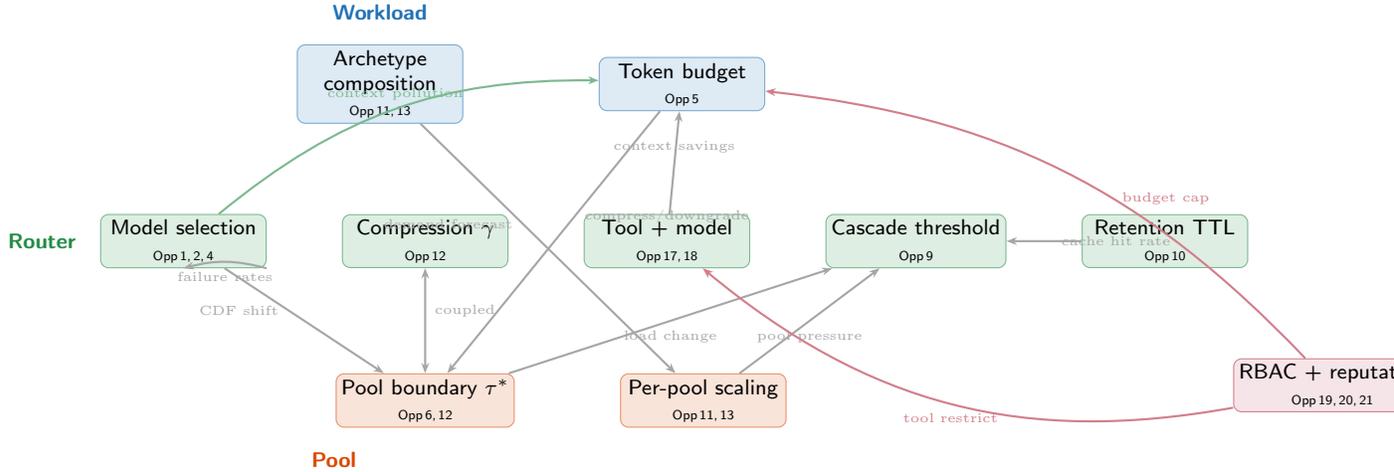

The opportunity is a \emph{meta-controller} that orchestrates the
individual adaptation loops as a unified system.  The architecture
follows the Monitor--Analyze--Plan--Execute (MAPE-K) reference
model from autonomic computing~\cite{kephart2003autonomic}:
\begin{enumerate}[nosep]
  \item \textbf{Monitor.}  Collect a per-session outcome record
        that unifies all signals: model sequence, tool outcomes,
        hallucination verdicts (HaluGate), safety scores
        (SafetyL1/L2), quality signal (Feedback Detector), turn
        count, token consumption, TTFT, context growth rate,
        failure-driven token waste (tokens consumed by turns that
        produced errors subsequently corrected or discarded), and
        task-completion indicator.  The cross-session failure
        memory---aggregated per (model, domain, failure
        class)---provides the fleet-wide baseline against which
        individual session outcomes are compared.
  \item \textbf{Analyze.}  Map each poor outcome to a failure class
        \emph{and} the knob that could have prevented it:
        ``12-turn session $\to$ wrong model at turn~2
        (\Cref{opp:session-cascade}),'' ``SLO miss $\to$ strong
        pool overloaded at cascade (\Cref{opp:pool-cascade}),''
        ``token explosion $\to$ no budget enforcement
        (\Cref{opp:token-budget}),'' ``40\% token waste $\to$
        hallucination at turn~3 polluted context for turns 4--8,
        domain-specific compression would have pruned the error
        trace (\Cref{opp:token-budget}).''  Cross-session
        failure schemata~\cite{correct2025} accelerate this
        analysis: structurally similar failures recur across
        sessions, so a new session's failure can be matched to a
        known schema rather than diagnosed from scratch.
  \item \textbf{Plan.}  Select knob adjustments that address the
        dominant failure classes while respecting interaction
        constraints: e.g., if model-selection changes shift the CDF
        enough to invalidate the current $\boundary^*$, schedule a
        joint $(\gamma, \boundary^*)$ re-optimization
        (\Cref{opp:workload-shaping}) in the same cycle.
  \item \textbf{Execute.}  Apply the new configuration via staged
        rollout: evaluate the candidate policy offline via
        importance-weighted off-policy
        estimation~\cite{deltope2024}, then route 5\% canary
        traffic under the new policy, measure the composite
        outcome, and promote to full traffic only if the outcome
        improves beyond a hysteresis threshold.
\end{enumerate}

\Cref{tab:mape} maps each opportunity to its MAPE-K role: the
observable it monitors, the knob it tunes, and the metric it
evaluates improvement against.

\begin{table}[t]
\centering
\caption{MAPE-K role of each opportunity: what the adaptation loop
monitors, which knob it adjusts, and how it evaluates improvement.}
\label{tab:mape}
\scriptsize
\renewcommand{\arraystretch}{1.10}
\begin{tabularx}{\textwidth}{c l X X X}
\toprule
\textbf{\#} & \textbf{Opportunity} & \textbf{Monitor (observable)}
  & \textbf{Knob (control)} & \textbf{Evaluate (metric)} \\
\midrule
\rowcolor{routergreen!6}
1  & Session-length RL
   & Session trajectories (model, tool, outcome per turn)
   & Model + tool selection policy
   & Median session length (turns, tokens) \\
2  & HaluGate reputation
   & Per-response hallucination verdicts
   & Per-(model, domain) routing weights
   & Hallucination rate per domain \\
\rowcolor{routergreen!6}
3  & Cumulative risk
   & Per-turn SafetyL1 confidence scores
   & Safety escalation threshold
   & FPR vs.\ attack catch rate \\
4  & Failure-class routing
   & Multi-signal outcome per (model, domain, failure class)
   & Failure-class routing weights
   & Oracle-utility capture rate \\
\rowcolor{routergreen!6}
5  & Token budget
   & Per-session token counter + failure-driven context bloat
   & Budget threshold + domain-specific compression policy
   & Token waste from failed sessions \\
\rowcolor{routergreen!6}
17 & Joint tool--model optimization
   & Per-(domain, turn, tool, model) outcomes + transition graph
   & Tool subset + resolving model + transition priors per dispatch
   & Aggregate session cost $\times$ task-completion rate \\
\rowcolor{routergreen!6}
18 & Gateway agent loops
   & Session depth, tool/cache/KV telemetry, cross-tenant outcomes
   & Instruction/tool surface + cache-block order + TTL/admission hints
   & A/B: tokens/rounds/TTFT vs.\ ITR-only or scheduler-only baselines \\
19 & RBAC enforcement
   & Bearer token JWT claims + model tool\_call outputs
   & Per-(role, tool, operation) permission matrix
   & Unauthorized tool-call block rate + false-denial rate \\
\rowcolor{routergreen!6}
20 & Caller reputation
   & Per-bearer-token security events across fleet
   & Reputation score $\to$ tool/model/budget restrictions
   & Cross-agent attack propagation prevention rate \\
21 & Behavioral commitments
   & Model boundary declarations in output text
   & Per-(token, agent) persistent block rules
   & Re-engagement-under-pressure prevention rate \\
\midrule
\rowcolor{poolorange!6}
6  & Follow-the-sun
   & Sliding-window prompt-length CDF
   & Pool boundary $\tau^*$
   & Fleet cost at SLO target \\
7  & Pool-state inference
   & Dispatch/completion counts, TTFT
   & Per-instance routing weights
   & Per-pool SLO attainment \\
\rowcolor{poolorange!6}
8  & Output-length routing
   & Actual completion lengths
   & Pool routing (prompt $+$ predicted output)
   & Short-pool overflow rate \\
9  & Pool-aware cascading
   & Estimated queue depth per pool
   & Cascade-abort threshold
   & Quality vs.\ TTFT SLO \\
\rowcolor{poolorange!6}
10 & KV-cache retention
   & Domain, turn number, OATS tool latency
   & Retention TTL + eviction priority
   & Cache hit rate on agent resume \\
\midrule
11 & Proactive scaling
   & Per-archetype traffic time series
   & Per-pool instance count
   & Forecast accuracy vs.\ aggregate \\
\rowcolor{gray!6}
12 & $(\gamma, \tau^*)$ co-adapt
   & Sliding-window CDF + current $(\gamma, \tau^*)$
   & Joint $(\gamma, \tau^*)$
   & Cost gap vs.\ boundary-only \\
13 & Mixed-archetype fleet
   & Router archetype breakdown
   & Fleet sizing weights
   & Complementarity savings \\
\rowcolor{gray!6}
14 & Energy routing
   & tok/W per pool configuration
   & Energy weight in decision rule
   & Fleet-wide tok/W at SLO \\
15 & Governance-as-code
   & Policy constraints + fleet topology
   & DSL rule set
   & Constraint satisfiability \\
\rowcolor{gray!6}
16 & Self-adaptation
   & All of the above (composite session + security record)
   & All knobs jointly (incl.\ RBAC, reputation, commitments)
   & Composite session outcome + security posture \\
\bottomrule
\end{tabularx}
\end{table}

The composite outcome function is the reward signal that ties the
system together.  It must weight multiple objectives---session
length, token cost, SLO attainment, failure rate, safety---into a
single scalar.  Recent work shows this is tractable:
PROTEUS~\cite{proteus2026} uses Lagrangian RL to accept accuracy
targets as runtime input without retraining;
BaRP~\cite{barp2025} trains under bandit feedback with
preference-tunable cost-quality tradeoffs, outperforming offline
routers by 12.5\%.  The meta-controller extends these ideas from
model selection to the full WRP control surface.

The router's structural advantage is again fleet-wide visibility:
it observes all sessions, all outcomes, and all knob settings
simultaneously.  No individual agent, model provider, or pool
instance has this view.  The meta-controller is the natural
convergence point where the individual adaptation loops become a
self-improving system.

\begin{frisk}
The MAPE-K pattern is mature in autonomic
computing~\cite{kephart2003autonomic} and has been applied to
database configuration tuning, where OnlineTune~\cite{onlinetune2022}
demonstrated safe online multi-knob adaptation in production cloud
databases (14--165\% improvement, 91--99.5\% fewer unsafe
recommendations) by embedding workload context and using contextual
Bayesian optimization with safe exploration.  The LLM inference
analogue replaces database knobs with WRP parameters and replaces
query throughput with the composite session-outcome metric.  The
research questions are: (a)~dimensionality---how many WRP knobs
can be jointly tuned before the search space becomes intractable
(OnlineTune handles $\sim$50 database knobs; WRP has $\sim$10,
well within range); (b)~evaluation cost---off-policy evaluation
from logged sessions avoids the need for live A/B tests on every
candidate configuration; (c)~interaction
strength---if the coupling between knobs is weak (e.g., changing
model selection does not materially shift the CDF), independent
loops suffice and the meta-controller adds unnecessary complexity.
FleetSim~\cite{fleetsim2026} can quantify the interaction strength
empirically.  The baseline is independent per-knob adaptation; the
metric is composite session outcome (quality-weighted session cost)
on a multi-week FleetSim trace with diurnal workload rotation.
\end{frisk}
\end{opportunity}

\section{Broader Context: Related Work}
\label{sec:related}

\paragraph{Surveys.}
Moslem and Kelleher~\cite{moslem2026routingsurvey} survey dynamic
model routing and cascading, covering query difficulty, human
preferences, clustering, uncertainty, RL, and multimodal routing;
their when/what/how taxonomy does not address pool architecture or
workload interactions.  Pan and Li~\cite{pan2025survey} cover the
full inference stack from operator optimization to serverless
deployment but treat routing as cluster-level load balancing.
Xia~\etal~\cite{xia2025taming} survey efficient LLM inference
serving with instance-level and cluster-level taxonomies.
None of these surveys formalize the cross-dimensional interactions
that the WRP framework targets.

\paragraph{Routing systems (beyond our project).}
RouteLLM~\cite{ong2025routellm} and Hybrid LLM~\cite{ding2024hybrid}
route between strong/weak models using preference data.
FrugalGPT~\cite{chen2023frugalgpt} and
AutoMix~\cite{madaan2024automix} use cascading.
Router-R1~\cite{routerr1_2025} and
R2-Router~\cite{r2router2026} apply RL-based selection.
MixLLM~\cite{lu2025mixllm} uses contextual bandits at NAACL 2025.
xRouter~\cite{xrouter2025} trains cost-aware RL routers over
20+ LLM tools with explicit cost-performance rewards.

\paragraph{Agent serving and failure analysis.}
AGSERVE~\cite{agserve2025} provides session-aware KV-cache
management and model cascading for agent workloads (NeurIPS 2025).
Helium~\cite{helium2026} treats agentic workflows as query plans
with proactive caching and cache-aware scheduling.
SUTRADHARA~\cite{sutradhara2026} co-designs the orchestrator with
the serving engine for tool-based agents.
Continuum~\cite{continuum2025} introduces KV-cache TTL pinning
for multi-turn agent scheduling on SWE-Bench and BFCL.
Concur~\cite{concur2026} applies agent-level admission control via
congestion-based concurrency regulation (ICML 2026).
A growing body of empirical work classifies agent failure
modes~\cite{agenticfaults2026} (37~fault types from 13,602
issues), model-specific failure signatures~\cite{erroratlas2026}
(83~models, 35~datasets), and multi-turn model-switch
degradation~\cite{handoffdrift2026} ($\pm$4--13\,pp from a
single handoff); none of these feed failure signals back into
routing decisions.

\paragraph{Fleet provisioning (beyond our project).}
M\'{e}lange~\cite{griggs2024melange} optimizes heterogeneous GPU
allocation.  SageServe~\cite{jia2025sageserve} integrates traffic
forecasting at Microsoft scale.

\paragraph{Disaggregated inference.}
Splitwise~\cite{patel2024splitwise},
DistServe~\cite{zhong2024distserve}, and
Sarathi-Serve~\cite{agrawal2024sarathi} disaggregate prefill and
decode.  Mooncake~\cite{qin2025mooncake} provides KV-cache-centric
disaggregation.  NVIDIA Dynamo~\cite{nvidia2026dynamo} enables
dynamic multi-node scheduling.

\paragraph{Energy efficiency (beyond our project).}
SweetSpot~\cite{patel2026sweetspot} provides analytical energy
prediction.  TokenPowerBench~\cite{niu2025tokenpowerbench} and
GreenServ~\cite{greenserv2026} benchmark and optimize
inference energy.

\section{Conclusion}
\label{sec:conclusion}

We have argued that the LLM inference optimization landscape is
usefully organized around three coupled dimensions---Workload, Router,
and Pool---whose interactions dominate real-world performance.  This
argument draws on a growing body of publications from the vLLM Semantic Router
project, covering signal composition, conflict detection, pool
routing, fleet provisioning, energy modeling, multimodal agent
routing, tool selection, security, semantic caching,
feedback-driven adaptation, hallucination detection,
low-latency embedding models, and hierarchical content-safety
classification for privacy and jailbreak protection.  Solving a problem in one dimension repeatedly exposed a dependency
on another.

The WRP matrix shows that the biggest open questions sit at the
intersections---particularly the sparse Router $\times$ Pool and
Pool $\times$ Router cells.  Our twenty-one proposed opportunities
(\Cref{tab:maturity}) address these gaps from the semantic router's
unique vantage point: the signal-decision architecture already
classifies every request by domain, modality, and complexity,
validates responses via HaluGate, tracks dispatch and completion
events per instance, and produces per-turn safety confidence
scores; these are untapped inputs to model selection, fleet
provisioning, cascading, and energy optimization.

Six are engineering-tier projects that can be built from existing
components: HaluGate-driven model reputation routing
(\Cref{opp:halugate-reputation}), runtime token-budget enforcement
(\Cref{opp:token-budget}), follow-the-sun pool rebalancing
(\Cref{opp:dynamic-boundary}), router-emitted KV-cache retention
directives (\Cref{opp:kv-retention}), governance-as-code
(\Cref{opp:governance}), and request-level RBAC enforcement
(\Cref{opp:rbac-enforcement}).  Fifteen are research-tier: offline RL
for session-length routing (\Cref{opp:session-cascade}), cumulative
multi-turn risk scoring (\Cref{opp:cumulative-risk}),
failure-class-aware routing from router outcome memory
(\Cref{opp:failure-routing}), fleet-learned joint tool--model
optimization (\Cref{opp:tool-pruning}), gateway-coordinated agent
loops for memory tiers and tool surfaces (\Cref{opp:gateway-agent-loops}),
fleet-wide caller reputation with cross-agent threat
propagation (\Cref{opp:caller-reputation}), router-enforced
behavioral commitments (\Cref{opp:behavioral-commitments}),
pool-state inference
(\Cref{opp:pool-feedback}), output-length-aware pool routing
(\Cref{opp:output-length}), pool-aware cascading
(\Cref{opp:pool-cascade}), online $(\gamma, \boundary^*)$
co-adaptation (\Cref{opp:workload-shaping}), mixed-archetype fleet
provisioning (\Cref{opp:mixed-archetype}), archetype-driven
proactive pool scaling (\Cref{opp:proactive-scaling}), energy
as a routing objective (\Cref{opp:energy-signal}), and closed-loop
self-adaptation across WRP knobs (\Cref{opp:self-adapt}).

As production workloads shift from chat toward agents and multimodal
applications, the cross-dimensional interactions identified here will grow in
importance.  We hope the WRP framework, including its assessment of
what is ready to build and what remains open, proves useful to
others working on these problems.

\bibliographystyle{unsrt}
\bibliography{references}

\end{document}